\documentclass{article}
\usepackage{ijcai22}

\usepackage{times}
\usepackage{soul}
\usepackage{url}
\usepackage[hidelinks]{hyperref}
\usepackage[utf8]{inputenc}
\usepackage[small]{caption}
\usepackage{graphicx}
\usepackage{amsmath}
\usepackage{amsthm}
\usepackage{booktabs}
\usepackage{algorithm}
\usepackage{algorithmic}
\usepackage{url}
\usepackage{bm}
\usepackage{amssymb,amsfonts,amsmath}
\usepackage{mathtools}
\usepackage{subfigure}
\usepackage{comment}
\usepackage{multirow}
\usepackage{enumitem}
\usepackage{pifont}
\newcommand\ourmethod{\textsc{HLT-MT}}

\newcommand{\cmark}{\ding{51}}

\title{\ourmethod{}: High-resource Language-specific Training for Multilingual \\ Neural Machine Translation}
\author{
  Jian Yang\textsuperscript{\rm 1 \thanks{\ Contribution during internship at Microsoft Research.}}, 
  Yuwei Yin\textsuperscript{\rm 2 *}, 
  Shuming Ma\textsuperscript{\rm 2}, 
  Dongdong Zhang\textsuperscript{\rm 2}, 
  Zhoujun Li\textsuperscript{\rm 1 \thanks{\ Corresponding author.}}, 
  Furu Wei\textsuperscript{\rm 2} 
  \affiliations
  \textsuperscript{\rm 1}State Key Lab of Software Development Environment, Beihang University
  \\ 
  \textsuperscript{\rm 2}Microsoft Research\\
  \{jiaya, lizj\}@buaa.edu.cn, \{v-yuweiyin, shumma, dozhang, fuwei\}@microsoft.com
}

\begin{document}

\maketitle

\begin{abstract}
Multilingual neural machine translation (MNMT) trained in multiple language pairs has attracted considerable attention due to fewer model parameters and lower training costs by sharing knowledge among multiple languages. Nonetheless, multilingual training is plagued by language interference degeneration in shared parameters because of the negative interference among different translation directions, especially on high-resource languages. In this paper, we propose the multilingual translation model with the high-resource language-specific training (\ourmethod{}) to alleviate the negative interference, which adopts the two-stage training with the language-specific selection mechanism. Specifically, we first train the multilingual model only with the high-resource pairs and select the language-specific modules at the top of the decoder to enhance the translation quality of high-resource directions. Next, the model is further trained on all available corpora to transfer knowledge from high-resource languages (HRLs) to low-resource languages (LRLs). Experimental results show that \ourmethod{} outperforms various strong baselines on WMT-10 and OPUS-100 benchmarks. Furthermore, the analytic experiments validate the effectiveness of our method in mitigating the negative interference in multilingual training.
\end{abstract}

\section{Introduction}
Recent advances in multilingual neural machine translation (MNMT) aim to build and deploy a single universal model in real industrial scenarios, which supports multiple translation directions by sharing model parameters \cite{shared_attention_mnmt,googlemnmt,mmnmt,m2m,mrasp}. Furthermore, parameter sharing across various languages encourages knowledge transfer, especially from the high-resource language (HRL) to low-resource language (LRL) and even enables zero-shot translation \cite{mmnmt,opus100}.

\begin{figure}[t]
\centering
\includegraphics[width=0.9\linewidth]{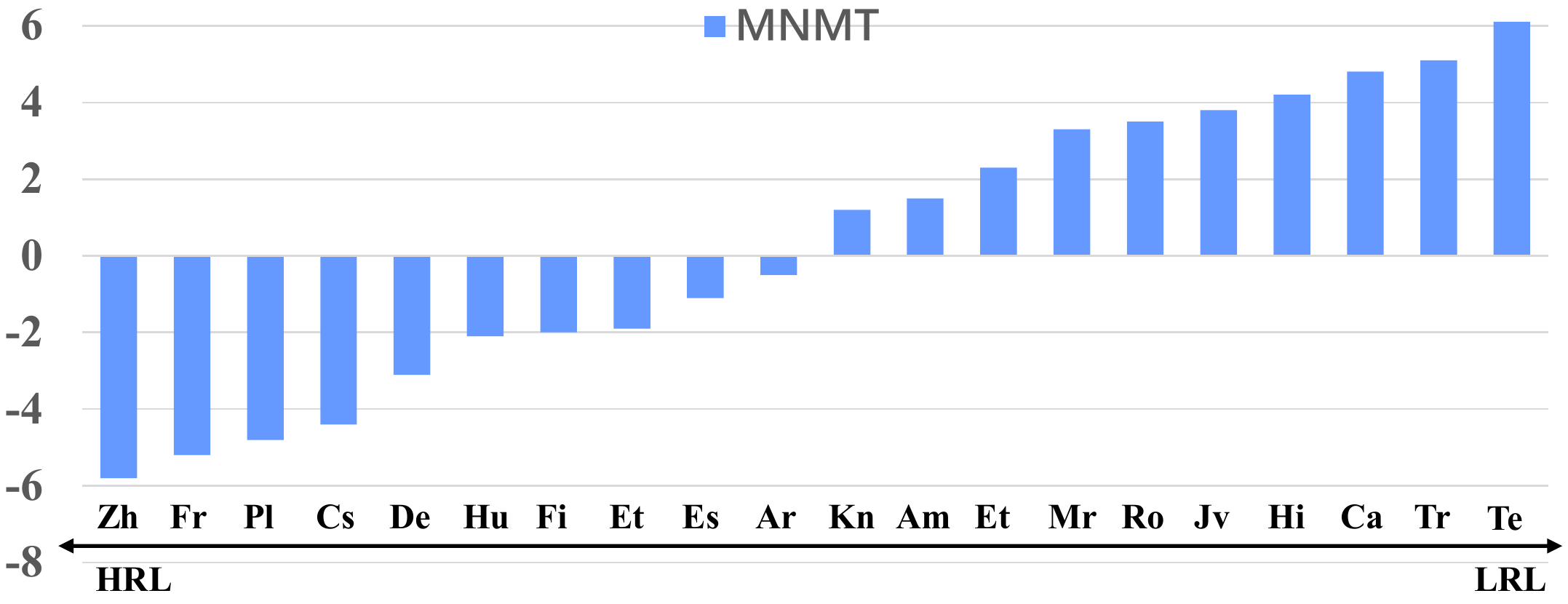}
\caption{Results of the multilingual translation model are reported as $\Delta$BLEU relative to the corresponding bilingual counterpart. The languages are arranged from high-resource languages (HRLs) to low-resource languages (LRLs).}
\vspace{-5pt}
\label{intro}
\end{figure}

While having attracted many interests, MNMT is still beleaguered by the \textit{negative language interference} residing in the multilingual parameter sharing \cite{xlmr,negative_interference_mnmt,adaptive_sparse_transformer}, where the multiple translation directions degrade performance on high-resource languages. In Figure \ref{intro}, the multilingual model outperforms the bilingual model on low-resource translations profited by knowledge transferability. But the multilingual model performs worse than the bilingual counterpart when it comes to high-resource translations, which is exactly the manifestation of the negative interference caused by the severe competition in multiple training directions among HRLs and LRLs.

To mitigate the negative interference, the previous works \cite{compact_adapter_mnmt,monolingual_adapter} propose the language adapter owing to its simplicity. The adapter modules are usually added to the encoder and decoder layers to indicate the source and target language. With the whole multilingual model frozen, only the parameters of the adapter continue to be tuned. But when the number of languages is large, the adapter methods suffer from the sharp increasing parameters and extra inference time. Another line of research is to extend the depth and width of the MNMT model and construct large-scale corpora to increase model capacity and include more languages \cite{opus100,deep_encoder_mnmt}. It still rises in substantial training and inference costs caused by numerous extra parameters.

%%%%%%%%%%%%%%%%%%%%%%%%%%%%%%%%%%%%%%%%%%%%%%%%%%%%%%%%%%%
\begin{figure*}[t]
\begin{center}
	\includegraphics[width=1.75\columnwidth]{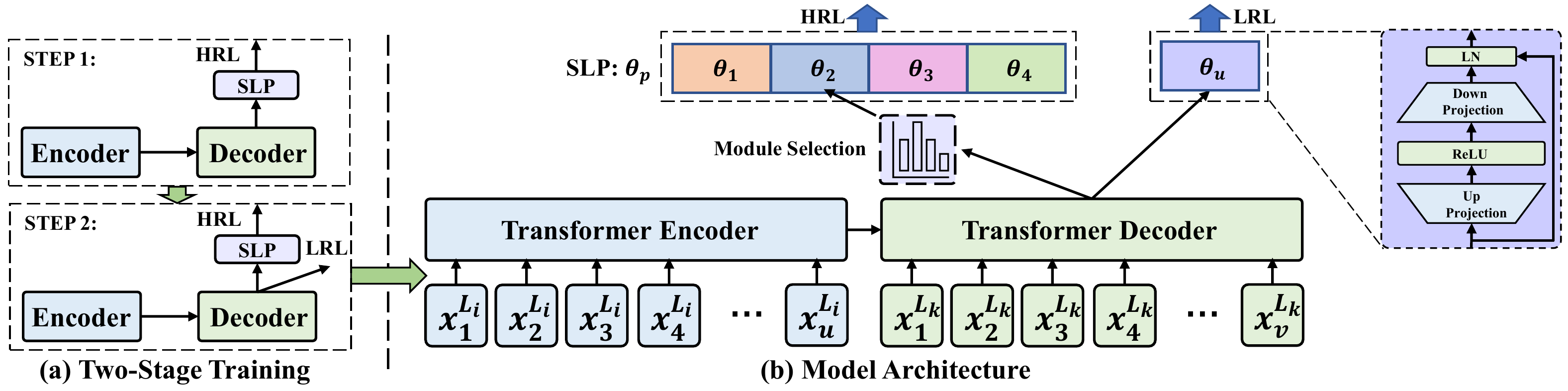}
	\caption{(a) is the two-stage training framework and (b) is the model architecture with the selective language-specific pool (SLP) for HRL and the universal layer for LRL. We first train the multilingual model on the high-resource pairs with the languages-specific pool and then continue tuning the model on all multilingual corpora to help transfer the knowledge from the high-resource to the low-resource languages. Given different translation directions of $K$ languages $L_{all}=\{L_{k}\}_{k=1}^{K}$, we first employ language-specific training for the high-resource translation directions and then continue to train on all available multilingual corpora. If $L_k$ is a HRL, we use the selection function $g(L_k)$ to decide which language-specific module will be used. Otherwise, we use the universal layer $\theta_{u}$ for the low-resource language $L_k$.}
	\label{model}
% 	\vspace{-10pt}
\end{center}
\end{figure*}
%%%%%%%%%%%%%%%%%%%%%%%%%%%%%%%%%%%%%%%%%%%%%%%%%%%%%%%%%%%

In this work, we propose a novel multilingual translation model with the high-resource language-specific training (HLT-MT) for one-to-many and many-to-many translation directions in a two-stage training framework. In the first stage, our model is trained only on high-resource languages with the language-specific modules to avoid the negative impact caused by LRLs. To address the negative interference problem among HRLs, we introduce a language-specific pool containing a sequence of independent modules for HRLs. Considering the increasing number of the languages, we apply the selection mechanism to the language-specific pool with the constrained size, denoted as selective language-specific pool (SLP), which enables different groups of certain languages to share the same module from SLP. After pretraining with the high-resource languages, we extract the shared representations of the decoder to transfer the fine-trained knowledge to the low-resource languages.

Our method aims to enhance translation quality of high-resource directions with the selective language-specific pool compared to the bilingual counterpart and then transfer the knowledge to the low-resource directions based on the bottom shared features. We conduct experiments on the WMT-10 benchmark of 11 languages and OPUS-100 benchmark of 95 languages. Experimental results demonstrate that our method significantly outperforms previous bilingual and multilingual baselines. Besides, extensive probing experiments are performed for the multilingual baseline and \ourmethod{}, helping further analyze how our method can benefit the multilingual machine translation. Empirical studies show that \ourmethod{} maintains a balance between language-agnostic and language-distinct features and thus helps to alleviate the negative language interference among various languages.

\section{Negative Language Interference}
\begin{figure}[t]
\centering
\includegraphics[width=0.7\columnwidth]{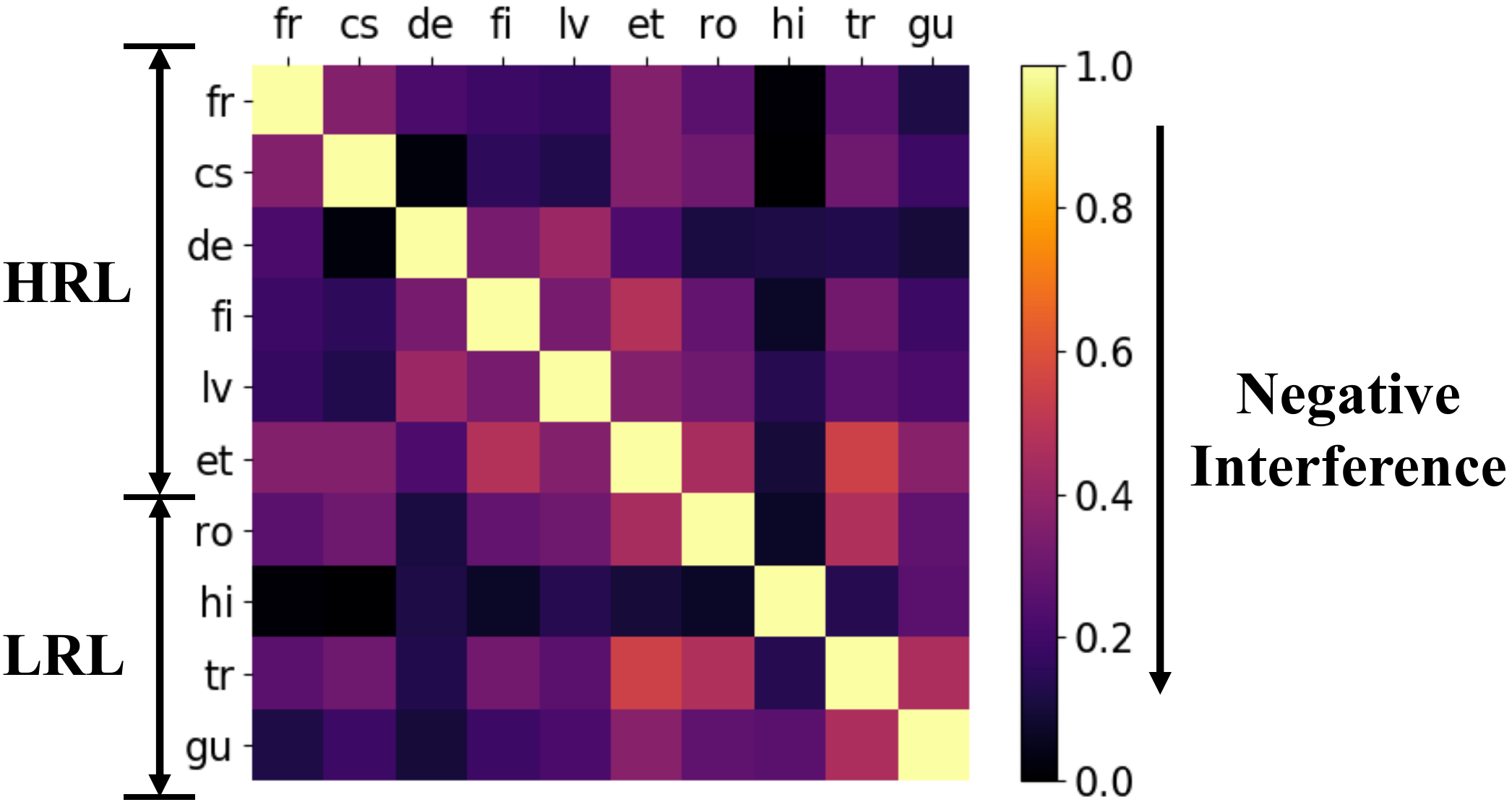}
\caption{Negative interference among multiple languages.}
\label{negative_inference}
% \vspace{-10pt}
\end{figure}

Multilingual translation model aims at transferring knowledge across languages to boost performance on low-resource languages, where the multilingual model is trained in multiple translation directions simultaneously to enable cross-lingual transfer through parameter sharing. However, different groups of languages have heterogeneous characteristics, such as different dictionaries and grammars. The previous works have shown \cite{negative_interference_mnmt,gradient_surgey} that knowledge transfer is not beneficial for all languages by sharing all parameters. To analyze the effect of mutual influence among different languages, we calculate the cosine similarities between gradients of two translation directions. In Figure \ref{negative_inference}, we observe that the certain high-resource language sustains negative interference from other HRLs and LRLs. For example, En$\to$Cs acutely conflicts with En$\to$De and En$\to$Hi in the second row of Figure \ref{negative_inference}.
This shows that HRLs are conducive to LRLs but may be hindered by other HRLs and LRLs in turn. To prevent the HRLs from negative interference introduced by LRLs, we focus on sufficiently training the high-resource directions and then continue tuning on all directions. To further address the conflicts among HRLs, we propose the selective language-specific pool for different high-resource languages.
Our method effectively mitigates the negative interference in the analytic experiments compared to baselines.

\section{Our Method}
In this section, we introduce \ourmethod{} for multilingual translation. We propose the two-stage training framework, where the selective languages-specific pool (SLP) and the universal layer are applied for HRLs and LRLs respectively.

\subsection{Overview of \ourmethod{}}
Given different translation directions of $K$ languages $L_{all}=\{L_{k}\}_{k=1}^{K}$, we employ SLP for high-resource directions and the universal layer for low-resource directions. 
Our method is illustrated in Figure \ref{model}, where the selective language-specific pool (SLP) for the high-resource languages is inserted into the top of the decoder denoted by $\theta_{p}=\{\theta_{t}\}_{t=1}^{T}$. 
$T$ is the constrained size of the selective language-specific pool and $T \leq K$ since the number of languages $K$ can be numerous. Thus, SLP contains $T$ individual modules with the same architecture, and each $\theta_{t}$ is a sub-network. If the translation direction $L_{i} \to L_{k}$ is the high-resource direction (from the source language $L_i$ to target language $L_k$), we only activate the relevant language-specific module from SLP for $L_k$. Otherwise, the universal layer is triggered for $L_k$.

\subsection{Multilingual Machine Translation}
Given the high-resource bilingual corpora $D^{h}=\{D_{m}^{h}\}_{m=1}^{M}$ and low-resource corpora $D^{l}=\{D_{n}^l\}_{n=1}^{N}$, where $M$ and $N$ separately represent the number of the high-resource and low-resource training corpora of $K$ languages $L_{all}=\{L_{k}\}_{k=1}^{K}$. The multilingual model is jointly trained on the union of the high- and low-resource training corpora $D^{h} \cup D^{l}$:
\begin{SmallEquation}
\begin{align}
\begin{split}
    \mathcal{L}_{MT} &=-\sum_{m=1}^{M} \mathbb{E}_{x,y \sim D_{m}^{h}} \left[ \log P(y|x; \Theta) \right] \\
                &-\sum_{n=1}^{N} \mathbb{E}_{x,y \sim D_{n}^{l}} \left[ \log P(y|x; \Theta) \right] 
    \label{objective-d}
\end{split}
\end{align}
\end{SmallEquation}where the first and second term denote objective of the high- and low-resource training corpora respectively. $\Theta$ are shared parameters for all languages. $x$ and $y$ are the sentence pair.

\subsection{High-resource Language-specific Training}
To prevent the high-resource languages from the negative interference caused by low-resource languages, we only train the model with SLP on high-resource directions, which effectively ameliorates translation quality of high-resource translation directions with slight extra parameters.

To take advantages of the cross-lingual pretrained encoder to boost model performance, our multilingual model is initialized by XLM-R \cite{xlmr}. Besides, we verify the effectiveness of our method on the Transformer model \cite{transformer} without any pretrained model. Given the source  sentence $x^{L_i}=\{x_1^{L_i},\dots,x_u^{L_i}\}$ with $u$ words and target sentence $x^{L_k}=\{x_1^{L_k},\dots,x_v^{L_k}\}$ with $v$ words, the shared features $h_{s}^{L_k}$ of the target language $L_k$ at the top of the decoder are obtained by the Transformer model:
\begin{SmallEquation}
\begin{align}
\begin{split}
    h_{s}^{L_k} = \text{Transformer}(x^{L_i}, x^{L_k};\Theta)
    \label{transformer}
\end{split}
\end{align}
\end{SmallEquation}where $h_s^{L_k}=\{h_{s_1}^{L_k},\dots,h_{s_j}^{L_k},\dots,h_{s_v}^{L_k}\}$ and $h_{s_j}^{L_k}$ is the $j$-th representation of the target token $h_j^{L_k}$ generated by the single shared Transformer encoder and decoder.

After obtaining a sequence of decoder representations $h_{s}^{L_k}$ of the high-resource language $L_k$, we project the language-agnostic representations to the language-distinct ones via the language-specific pool with the selective mechanism. Given the translation direction $L_i \to L_k$ ($1 \leq i,k \leq K \land i \neq k$) and the selective language-specific pool $\theta_{p} = \{\theta_{t}\}_{t=1}^{T}$, the corresponding module $\theta_{g(L_{i},L_{k})}$ is used to generate the language-distinct representations. $g(\cdot)$ is a map function that maps the language index to the corresponding module index: $L_k \in \{1,\dots,
K\} \longmapsto t \in \{1,\dots,T\}$. $g(L_{i},L_{k})$ is the map function only depending on the target language and thus can be simplified into $g(L_{k})$. Therefore, we project the language-agnostic representations $h_s^{L_k}$ to the language-specific features $h_{b}^{L_k}$ using function $\mathcal{F}_{\theta_{g(L_k)}}$ as below:
\begin{SmallEquation}
\begin{align}
\begin{split}
    h_{b}^{L_k} = \mathcal{F}_{\theta_{g(L_k)}}(h_{s}^{L_k})
    \label{hard_adapter}
\end{split}
\end{align}
\end{SmallEquation}where $h_{s}^{L_k}$ is the representations generated by the shared parameters. $\mathcal{F}_{\theta_{g(L_k)}}$ is a function defined as below:
\begin{SmallEquation}
\begin{align}
\begin{split}
    \mathcal{F}_{\theta_{g(L_k)}}(h_{s}^{L_k}) = f(W^d_{{g(L_k)}}\sigma (W^u_{{g(L_k)}}h_s^{L_k}) + h_s^{L_k})
    \label{adapter}
\end{split}
\end{align}
\end{SmallEquation}where $f(\cdot)$ is the layer normalization and $\sigma(\cdot)$ is the ReLU activation function. $W^u_{{g(L_k)}} \in R^{d_e \times d_h}$ is the up-projection matrix and $W^d_{{g(L_k)}} \in R^{d_h \times d_e}$ is the down-projection matrix as shown in the right part of Figure \ref{model}, where $d_e$ and $d_h$ are the embedding size and hidden size of SLP ($d_e < d_h$). 

Another issue is how to design a proper map function $g(\cdot)$ with an appropriate selection mechanism for the translation direction $L_i \to L_k$. In our work, each source sequence is prefixed with a special target language symbol to indicate the translation direction, which enables the decoder to correctly generate the target sentence with the shared decoder parameters. Therefore, the embedding of the target language symbol is used to select the language-specific module from SLP. The selection function $g(\cdot)$ is defined as:
\begin{SmallEquation}
\begin{align}
\begin{split}
    g(L_k) = \mathop{\arg\max}_{1 \leq t \leq T}  \frac{exp(e^{L_k}_t)}{\sum_{i=1}^{T} exp(e^{L_k}_i)}
    \label{select_hard_adapter}
\end{split}
\end{align}
\end{SmallEquation}where $e^{L_k}=W_{g}E[L_{k}]$ of $T$ dimensions. $E[L_k]$ denotes the embedding of the target language $L_k$ symbol. $W_g \in R^{d_e \times T}$ maps the target embedding to the vector $e^{L_k}$, where $e^{L_k}_i$ is the $i$-th element of $e^{L_k}$ and SLP is comprised of $T$ sub-networks. The sub-network with the highest probability will be selected to produce the language-specific features.

Equation \ref{hard_adapter} and \ref{select_hard_adapter} are only related to $\theta_{g(L_k)}$ and thus can not propagate gradients to all language-specific parameters. The selective language-specific pool $\theta_{p}=\{\theta_t\}_{t=1}^{T}$ contains a set of modules described in Equation \ref{adapter}. To tackle the undifferentiable problem of SLP, we use the weighted average to ensure gradients to be propagated to all language-specific modules:
\begin{SmallEquation}
\begin{align}
\begin{split}
     h_{b}^{L_k} = \sum_{t=1}^{T}\alpha_t^{L_k} \mathcal{F}_{\theta_{t}}(h_{s}^{L_k}) 
    \label{soft_adapter}
\end{split}
\end{align}
\end{SmallEquation}where $\alpha_{t}^{L_k}$ is calculated by the target embedding and softmax function. We project the target embedding $E[L_{k}]$ the probability vector $e^{L_k}=W_{g}E[L_{k}]$ with the learned matrix $W_{g}$.
\begin{SmallEquation}
\begin{align}
\begin{split}
    % \alpha_t^{L_k} = exp(e^{L_k}_t)/\sum_{i=1}^{T} exp(e^{L_k}_i)
    \alpha_t^{L_k} = \frac{exp(e^{L_k}_t)}{\sum_{i=1}^{T} exp(e^{L_k}_i)}
    \label{soft_adapter_probabilities}
\end{split}
\end{align}
\end{SmallEquation}where $e^{L_k}=W_{g}E[L_{k}]$ of $T$ dimensions. $W_g \in R^{d_e \times T}$ and $E[L_k]$ denote the language embedding of $L_k$. $d_e$ is the embedding size. $\alpha^{L_k}_t$ is the $t$-th element of the vector $\alpha^{L_k}$.

In the practice training, we alternately leverage the Equation \ref{hard_adapter} and \ref{soft_adapter} with equal probabilities to learn the map function and generate the language-distinct representations $h_b^{L_k}$. Finally, the representations are used to generate the target sentence $x_{L_k}$:
\begin{SmallEquation}
\begin{align}
\begin{split}
    x^{L_k} = softmax(W^{o} h_{b}^{L_k})
    \label{generation}
\end{split}
\end{align}
\end{SmallEquation}where $x^{L_k}$ is the target sentence and $W^o \in d_e \times V$ is the output matrix, where $V$ is the vocabulary size.

\subsection{Low-resource Transfer}
After training the multilingual model on the high-resource language pairs, the bottom features generated by the shared parameters are utilized for the low-resource target sentence generation. Then, our multilingual model is jointly trained on the high-resource bilingual corpora $D^h$ and low-resource bilingual corpora $D^l$.
Given the low-resource translation direction $L_i \to L_j$, the shared representations $h^{L_j}_s$ generated by the shared parameters $\Theta$ are fed into a universal layer $\theta_u$:
\begin{SmallEquation}
\begin{align}
\begin{split}
    h_{b}^{L_j} = \mathcal{F}_{\theta_{u}}(h_{s}^{L_j})
    \label{universal_adapter}
\end{split}
\end{align}
\end{SmallEquation}where $h_{b}^{L_j}$ are features generated by the shared parameters in Equation \ref{transformer}. $\theta_{u}$ is a sub-network same as the $\theta_t$ ($1 \leq t \leq T$) of SLP $\theta_p=\{\theta_t\}_{t=1}^{T}$. All low-resource languages share the same universal layer to project the shared features $h_s^{L_j}$ to the last representations $h_b^{L_j}$. Then the target sentence is produced by $h_b^{L_j}$ and output matrix $W^o$ similar to Equation \ref{generation}.

\subsection{Training Strategy}
Our model first accumulates the cross-entropy loss on the high-resource pairs and the disparity loss. Then, the multilingual model trained on multilingual corpora to maintain the performance of high-resource languages and meanwhile transfer the knowledge to the low-resource languages.

\paragraph{Disparity Loss} To encourage different languages to select different language-specific modules of SLP, we minimize the disparity loss $L_d$, which measures the similarity of language-specific layer selection among languages.
\begin{SmallEquation}
\begin{align}
\begin{split}
    % \mathcal{L}_d = \sum_{i=1}^{N}\sum_{j=i+1}^{N} (\alpha^{L_i} \cdot \alpha^{L_k})
    \mathcal{L}_d = \sum_{i=1}^{M}\sum_{k=i+1}^{M} (\alpha^{L_i} \cdot \alpha^{L_k})
    \label{disparsity_loss}
\end{split}
\end{align}
\end{SmallEquation}where $\alpha^{L_i}, \alpha^{L_k} \in R^{T}$ denote the selection probabilities generated by Equation \ref{soft_adapter_probabilities}, where SLP contains $T$ modules. 

\paragraph{High-resource Language-specific Training}
The objective is to minimize the cross-entropy loss of high-resource training corpora and the auxiliary disparity loss jointly as below:
\begin{SmallEquation}
\begin{align}
\begin{split}
     \mathcal{L}_{high} = -\sum_{m=1}^{M} \mathbb{E}_{x,y \sim D_{m}^{h}} \left[ \log P(y|x;\Theta,\theta_{p}) \right] + \mathcal{L}_d
     \label{high_resource_loss}
\end{split}
\end{align}
\end{SmallEquation}where $\Theta$ denotes all shared parameters and $\theta_p$ are parameters of SLP. $x$ and $y$ are sentence pair.

\paragraph{Multilingual Training} After trained on the high-resource directions, the multilingual model is continued to be tuned on the union of the high- and low-resource corpora $D_h \cup D_l$ with the extra SLP for high-resource languages and the universal layer for low-resource languages:
\begin{SmallEquation}
\begin{align}
\begin{split}
     \mathcal{L}_{all} &= -\sum_{m=1}^{M} \mathbb{E}_{x,y \sim D_{m}^{h}} \left[ \log P(y|x; \Theta, \theta_{p}) \right] \\
     &-\sum_{n=1}^{N} \mathbb{E}_{x,y \sim D_{n}^{l}} \left[ \log P(y|x;\Theta, \theta_{u}) \right]
     \label{all_loss}
\end{split}
\end{align}
\end{SmallEquation}where $\Theta$ are shared parameters for all languages. SLP contains a list of language-specific layers for HRLs and $\theta_u$ is a universal layer for LRLs.

\section{Experiments}

\subsection{Datasets}
To evaluate our method, we conduct experiments on the WMT-10 and the OPUS-100 dataset.
\paragraph{WMT-10}
We use a collection of parallel data in different languages from the WMT datasets to evaluate the models \cite{zcode}, The parallel data is between English and other 10 languages, including French (Fr), Czech (Cs), German (De), Finnish (Fi), Latvian (Lv), Estonian (Et), Romanian (Ro), Hindi (Hi), Turkish (Tr) and Gujarati (Gu). 
\paragraph{OPUS-100}
We use the OPUS-100 corpus \cite{opus100} for massively multilingual machine translation. 
OPUS-100 is an English-centric multilingual corpus covering 100 languages, which is randomly sampled from the OPUS collection. We obtain 94 English-centric language pairs after dropping out 5 languages, which lack corresponding test sets.

\subsection{Baselines}
Our method is compared to the bilingual and multilingual methods. For a fair comparison, \textbf{XLM-R} and \textbf{LS-MNMT} are initialized by XLM-R \cite{xlmr}. \textbf{BiNMT} \cite{transformer} is the bilingual Transformer model. \textbf{MNMT} \cite{googlemnmt} is jointly trained on all directions, where the target language symbol is prefixed to the input sentence. \textbf{mBART} \cite{mbart} is an encoder-decoder pretrained model and then is finetuned on all corpora. \textbf{XLM-R} \cite{xlmr} is initialized by the pretrained model XLM-R \cite{xlmt}. \textbf{LS-MNMT} \cite{m2m} integrates the language-specific layers of all languages into the end of the decoder.

\subsection{Training and Evaluation}
We adopt Transformer as the backbone model for all experiments. We train multilingual models with Adam ($\beta_{1}=0.9$, $\beta_{2}=0.98$). The learning rate is set as 5e-4 with a warm-up step of 4,000. The models are trained with the label smoothing cross-entropy with a smoothing ratio of 0.1. The batch size is 4096 tokens on 64 Tesla V100 GPUs. For WMT-10, we first train the multilingual model with 6 languages and then finetunes on all languages. For OPUS-100, the model is trained in the languages where the number of pairs exceeds 10K. The evaluation metric is the case-sensitive detokenized sacreBLEU \cite{sacrebleu}.

%%%%%%%%%%%%%%%%%%%%%%%%%%%%%%%%%%%%%%%%%%%%%%%%%%%%%%%%%%%%%%%%%%%%
\begin{table*}[t]
\centering
%\resizebox{0.75\textwidth}{!}{
\resizebox{0.85\textwidth}{!}{
\begin{tabular}{l|l|c|cccccccccc|c}
\toprule
\multicolumn{2}{l|}{En$\rightarrow$X test sets} & \#Params & Fr & Cs & De & Fi & Lv & Et & Ro & Hi & Tr & Gu & Avg$_{all}$ \\
\midrule
1$\rightarrow$1 & BiNMT \cite{transformer} &242M/10M& 36.3 & 22.3 & 40.2 & 15.2 & 16.5 & 15.0 & 23.0 & 12.2 & 13.3 & 7.9 & 20.2\\
\midrule
\multirow{5}{*}{1$\rightarrow$N} & MNMT \cite{googlemnmt}  &242M& 34.2 & 20.9 & 40.0 & 15.0 & 18.1 & 20.9 & 26.0 & 14.5 & 17.3 & 13.2 & 22.0 \\
& mBART~\cite{mbart} &611M& 33.7 & 20.8 & 38.9 & 14.5 & 18.2 & 20.5 & 26.0 & 15.3 & 16.8 & 12.9 & 21.8 \\
& XLM-R \cite{xlmr} &362M& 34.7 &  21.5 &  40.1 &  15.2 & 18.6 &  20.8 &  26.4 & 15.6 &  \bf 17.4 & \bf 14.9 & 22.5 \\
& LS-MNMT \cite{m2m}  &409M  &35.0  &21.7  &40.6   &15.5 &18.9  &21.0 &26.2  &14.8 &16.5  &12.8 &22.3  \\
& \bf \ourmethod{} (Our method)  & 381M &  \bf 36.2 & \bf 22.2 & \bf 41.8 & \bf 16.6 & \bf 19.5 & \bf 21.1 & \bf 26.6 & \bf 15.8 & 17.1 & 14.6 & \bf 23.2 \\
\midrule
\multirow{5}{*}{N$\rightarrow$N} & MNMT \cite{googlemnmt} & 242M & 34.2 & 21.0 & 39.4 & 15.2 & 18.6 & 20.4 & 26.1 & 15.1 & 17.2 & 13.1 & 22.0 \\
& mBART~\cite{mbart} & 611M & 32.4 & 19.0 & 37.0 & 13.2 & 17.0 & 19.5 & 25.1 & \bf 15.7 & 16.7 & 14.2 & 21.0 \\
& XLM-R \cite{xlmr} &362M& 34.2 &  21.4 &  39.7 &  15.3 & 18.9 &  20.6 &  26.5 & 15.6 &  17.5 & 14.5 & 22.4 \\
& LS-MNMT \cite{m2m}  &409M&  34.8 & 21.1 & 39.3 & 15.2 & 18.7 & 20.5 & 26.3 & 14.9 & 17.3 & 12.3 & 22.0 \\
& \bf \ourmethod{} (Our method) &381M& \bf 35.8 & \bf 22.4 & \bf 41.5 & \bf 16.3 & \bf 19.6 & \bf 21.0 & \bf 26.6 & \bf 15.7 & \bf 17.6 & \bf 14.7 & \bf 23.1 \\
\bottomrule
\end{tabular}}
\caption{En$\rightarrow$X evaluation results for bilingual (1$\rightarrow$1), one-to-many (1$\rightarrow$N), and many-to-many (N$\rightarrow$N) models on WMT-10. The languages are ordered from high-resource languages (left) to low-resource languages (right).}
\label{table:wmt10}
% \vspace{-5pt}
\end{table*}
%%%%%%%%%%%%%%%%%%%%%%%%%%%%%%%%%%%%%%%%%%%%%%%%%%%%%%%%%%%%%%%%%%%%

%%%%%%%%%%%%%%%%%%%%%%%%%%%%%%%%%%%%%%%%%%%%%%%%%%%%%%%
\begin{table*}[t]
\centering
%\resizebox{0.75\textwidth}{!}{
\resizebox{0.85\textwidth}{!}{
\begin{tabular}{l|c|ccccc|ccccc}
\toprule
\multirow{2}{*}{Models (N$\rightarrow$N)} &\multirow{2}{*}{\#Params} & \multicolumn{5}{c|}{X$\rightarrow$En} &\multicolumn{5}{c}{En$\rightarrow$X} \\ \cmidrule{3-12}
&  & High$_{45}$ & Med$_{21}$ & Low$_{28}$ & Avg$_{94}$ & WR & High$_{45}$ & Med$_{21}$ & Low$_{28}$ & Avg$_{94}$ & WR \\\midrule
Previous Best System \cite{opus100}             &254M  &30.3 &32.6 &31.9  &31.4 & -&23.7 &25.6 &22.2 &24.0 & -\\\midrule
MNMT \cite{googlemnmt}                          &242M  &32.3 &35.1 &35.8  &33.9 & \textit{ref} &26.3 &31.4 &31.2 &28.9 &\textit{ref} \\
XLM-R \cite{xlmr}                               &362M  &33.1 &35.7 &36.1  &34.6 & -&26.9 &31.9 &31.7 &29.4 &- \\
LS-MNMT \cite{m2m}                              &456M  &33.4 &35.8 &35.9  &34.7 & -&27.5 &31.6 &31.5 &29.6 &- \\
\bf \ourmethod{} (Our method)                   &391M& \bf 34.1  &\bf 36.6 &\bf 36.1  &\bf 35.3 &72.3 &\bf 27.6 &\bf 33.3 &\bf 31.8 &\bf 30.1 &77.7  \\
\bottomrule
\end{tabular}}
\caption{X$\rightarrow$En and En$\rightarrow$X test BLEU for high/medium/low resource language pairs in many-to-many setting on OPUS-100 test sets. The BLEU scores are average across all language pairs in the respective groups. ``WR'': win ratio (\%) compared to \textit{ref} (MNMT).}
\label{table:opus}
% \vspace{-5pt}
\end{table*}
%%%%%%%%%%%%%%%%%%%%%%%%%%%%%%%%%%%%%%%%%%%%%%%%%%%%%%%

\subsection{Main Results}

\paragraph{WMT-10} As shown in Table \ref{table:wmt10}, our method clearly improves multilingual baselines by a large margin in 10 translation directions. Previously, multilingual machine translation underperforms the bilingual translation model in rich-resource scenarios. It is worth noting that our multilingual machine translation baseline XLM-R is already very competitive initialized by the cross-lingual pretrained model. Interestingly, our method outperforms the bilingual baseline in the high-resource translation direction, such as En$\to$De translation direction (+1.5 BLEU points).
Our method consistently outperforms the multilingual baseline on all language pairs, confirming that using \ourmethod{} to alleviate negative interference can help boost performance.

\paragraph{OPUS-100} In Table \ref{table:opus}, we observe that \ourmethod{} achieves reasonable results on 94 translation directions. The improvement can be attributed to the high-resource training with the selective language-specific pool, which avoids the competition between high-resource and low-resource training directions. Another benefit of our approach is light and convenient to be applied to different backbone models since the parameters of the selective language-specific pool on the top of the decoder are tiny compared to all parameters.

\section{Analysis}
\label{section_analysis}

%%%%%%%%%%%%%%%%%%%%%%%%%%%%%%%%%%%%%%%%%%%%%%%%%%%%%%%%%%%%
\begin{figure}[t]
    \centering
    \subfigure[Number of Module]{
    \includegraphics[width=0.4\columnwidth]{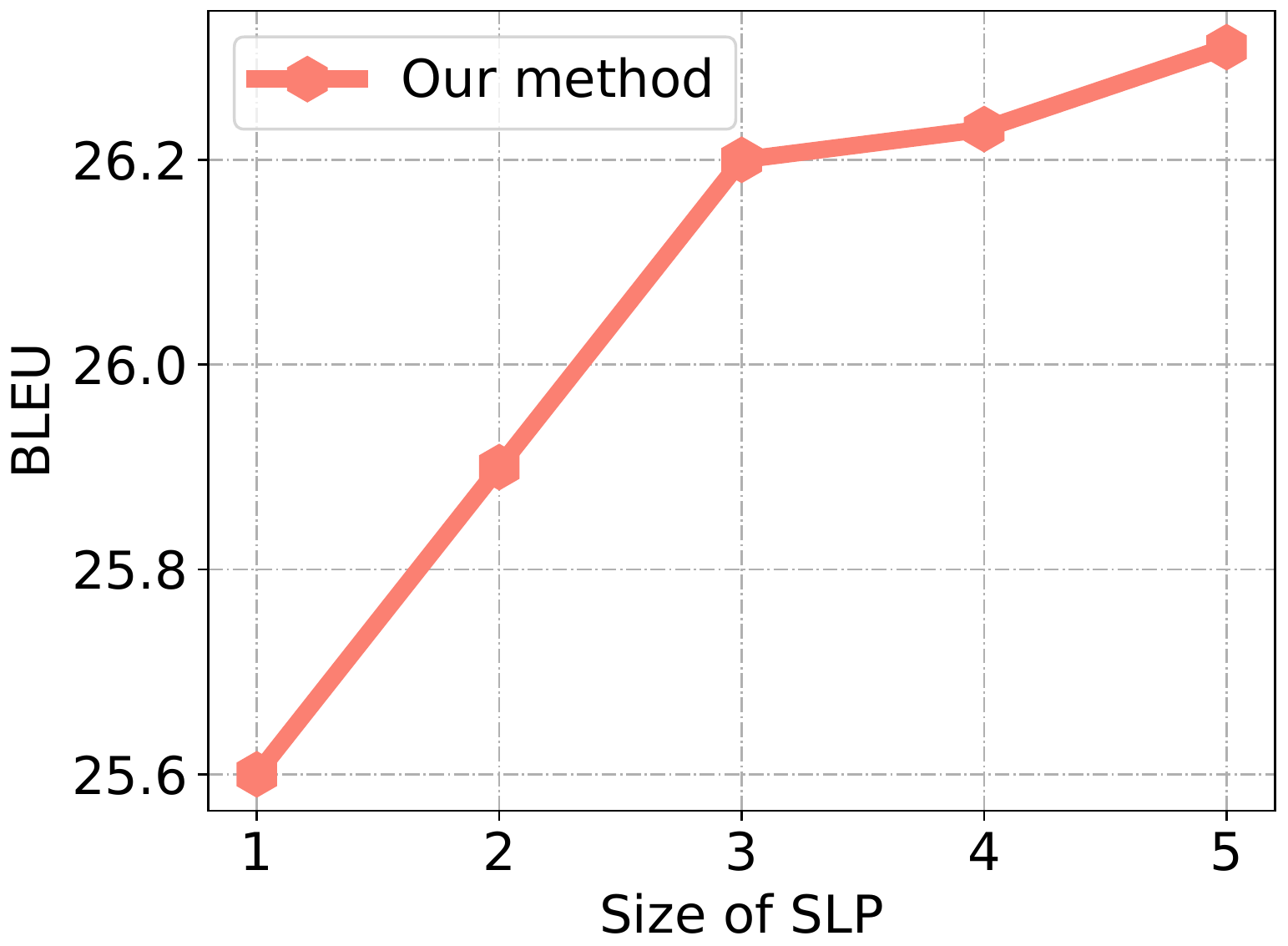}
    \label{slp_size}
    }
    \subfigure[Hidden size]{
    \includegraphics[width=0.4\columnwidth]{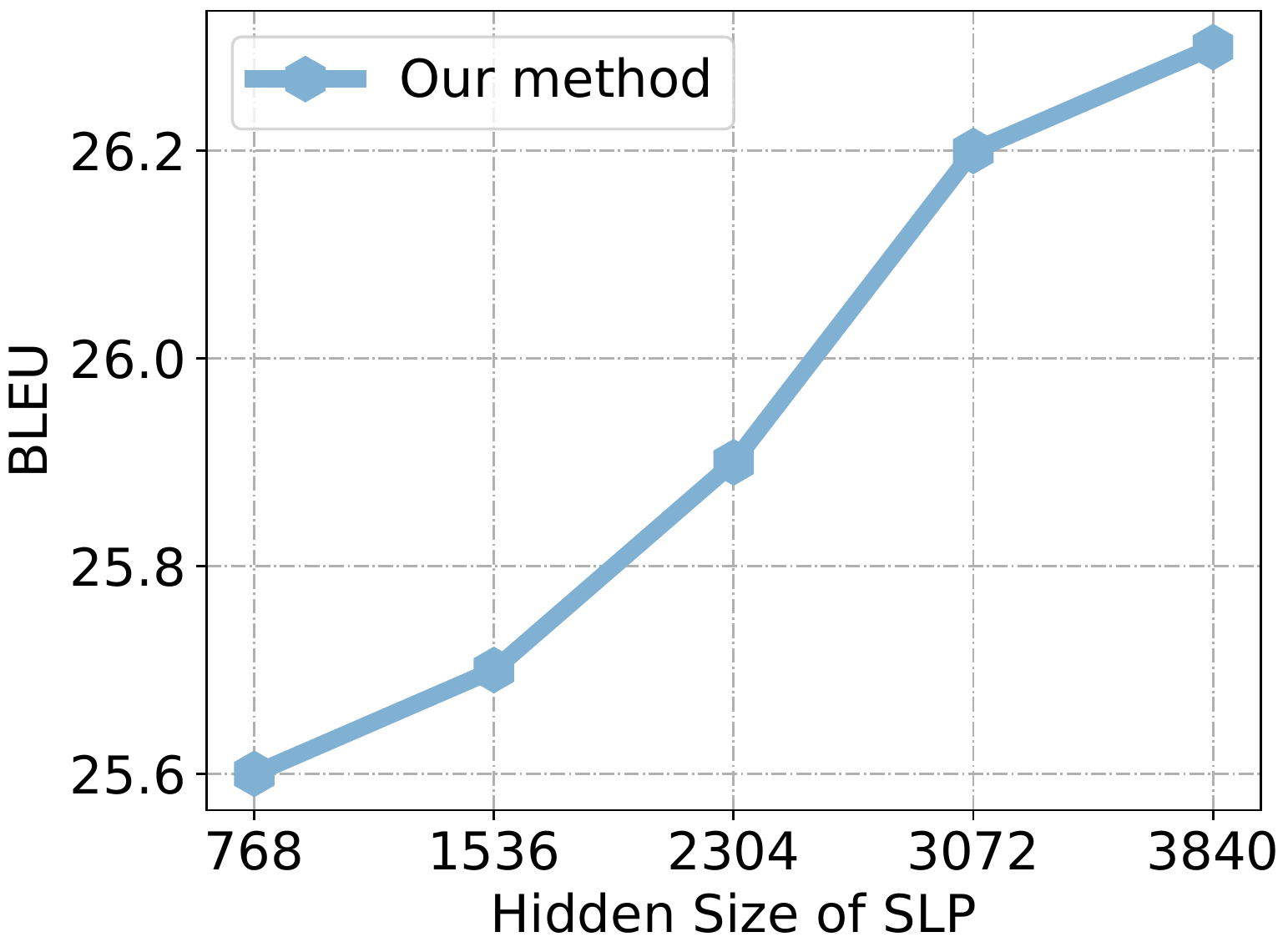}
    \label{slp_dim}
    }
    \caption{Average results of En$\to$X high-resource directions (Fr, Cs, De, Fi, Lv, and Et) on the WMT-10 benchmark.} 
    \vspace{-10pt}
\end{figure}
%%%%%%%%%%%%%%%%%%%%%%%%%%%%%%%%%%%%%%%%%%%%%%%%%%%%%%%%%%%%

\paragraph{Size of Language-specific Parameters} The size of the selective language-specific pool depends on the two key factors, namely the hidden size $d_h$ and the number of modules $T$ described in Equation \ref{adapter} and \ref{hard_adapter}. We tune the different values of $d_h$ and $T$ in Figure \ref{slp_size} and \ref{slp_dim} on the WMT-10 dataset. Naturally, the selective language-specific pool with a larger capacity leads to better performance. Increasing the number of the selective pool brings more improvement than the improvement of hidden size. Our method can efficiently reduce the language-specific parameters ($T=3$) using the selection mechanism and get comparable results compared to the baseline, where each high-resource language has the independent language-specific layer ($T=6$).

%%%%%%%%%%%%%%%%%%%%%%%%%%%%%%%%%%%%%%%%%%%%%%%%%%%%%%%%%%%%%%%%%
\begin{table}[t]
\centering
\resizebox{0.75\columnwidth}{!}{
\begin{tabular}{c|cc|ccc}
\toprule
  XLM-R   &  Two-stage Training  &  SLP  &  Avg$_{high}$ &  Avg$_{low}$ &  Avg$_{all}$ \\
\midrule
       &        &        &24.9 &17.8 & 22.0 \\
       & \cmark &        &25.4 &18.0 & 22.4 \\
       & \cmark & \cmark &26.0 &18.1 & 22.8 \\
\cmark &        &        &25.2 &18.5 & 22.5  \\
\cmark & \cmark &        &26.0 &17.9 & 22.8 \\
\cmark & \cmark & \cmark &\bf 26.2 &\bf 18.5 &\bf 23.2 \\
\bottomrule
\end{tabular}
}
\caption{Ablation study of our proposed approach on the WMT-10 benchmark. Our method can be easily initialized by the cross-lingual pretrained model XLM-R to enhance the performance.}
\label{ablation_study}
% \vspace{-10pt}
\end{table}
%%%%%%%%%%%%%%%%%%%%%%%%%%%%%%%%%%%%%%%%%%%%%%%%%%%%%%%%%%%%%%%%%%%%%%%%%%%%

\paragraph{Ablation Study} 
In Table \ref{ablation_study}, we empirically validate our approach on the different backbone models including Transformer \cite{transformer} without any pretrained model and XLM-T \cite{xlmt} initialized by the cross-lingual pretrained model XLM-R \cite{xlmr}. High-resource training significantly helps improve the model performance but has trouble in effectively handling the low-resource translation directions merely by sharing all parameters, which is caused by the competition in the shared parameters between high-resource and low-resource languages. By introducing the selective language-specific pool (SLP) and extracting the bottom representations for low-resource languages, our approach ameliorates all translation directions.

%%%%%%%%%%%%%%%%%%%%%%%%%%%%%%%%%%%%%%%%%%%%%%%%%%%%%%%%%%%%
\begin{figure}[t]
    \centering
    \subfigure[Baseline]{
    \includegraphics[width=0.38\columnwidth]{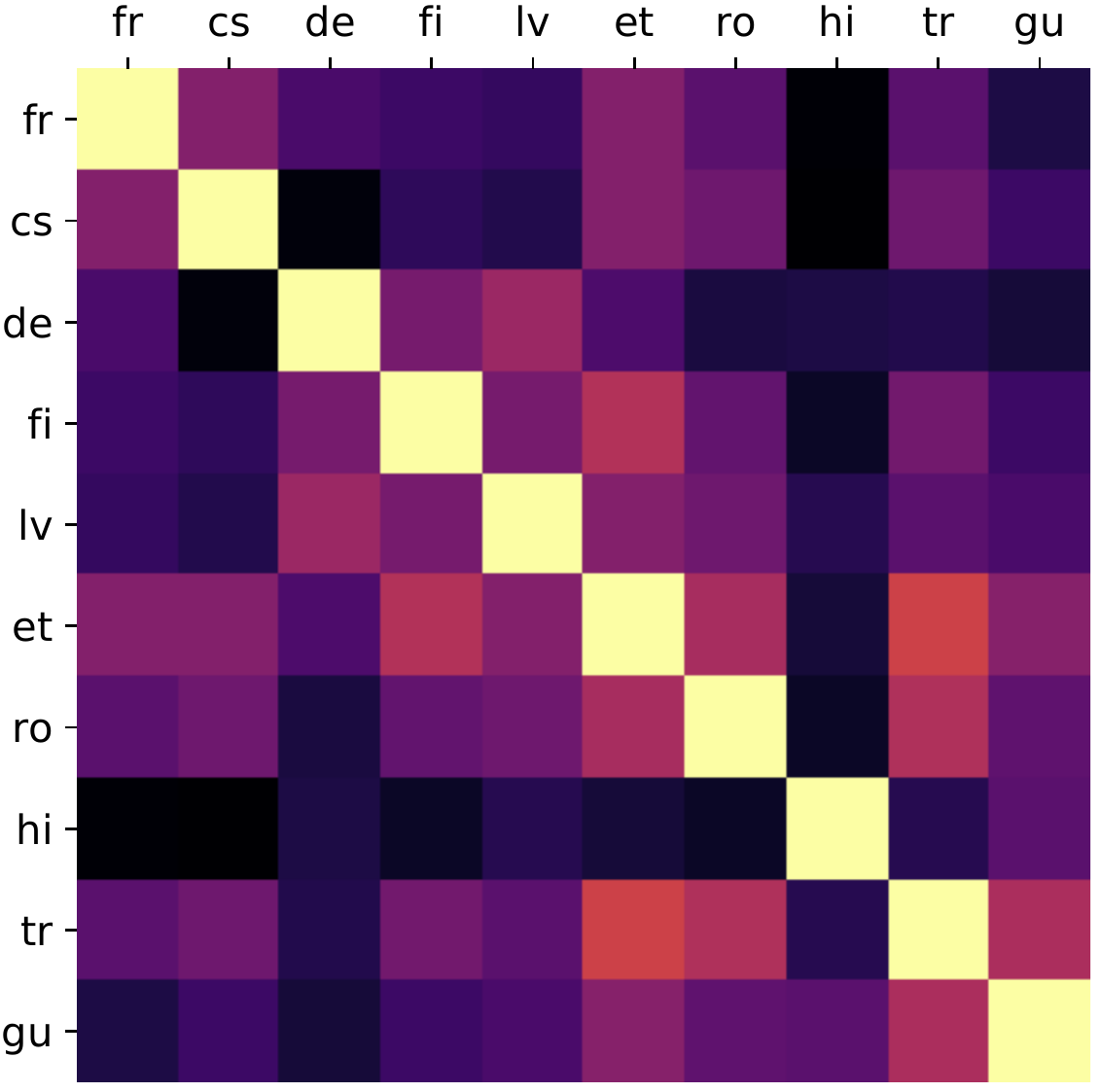}
    \label{grads_conflict_our}
    }
    \subfigure[Our method]{
    \includegraphics[width=0.38\columnwidth]{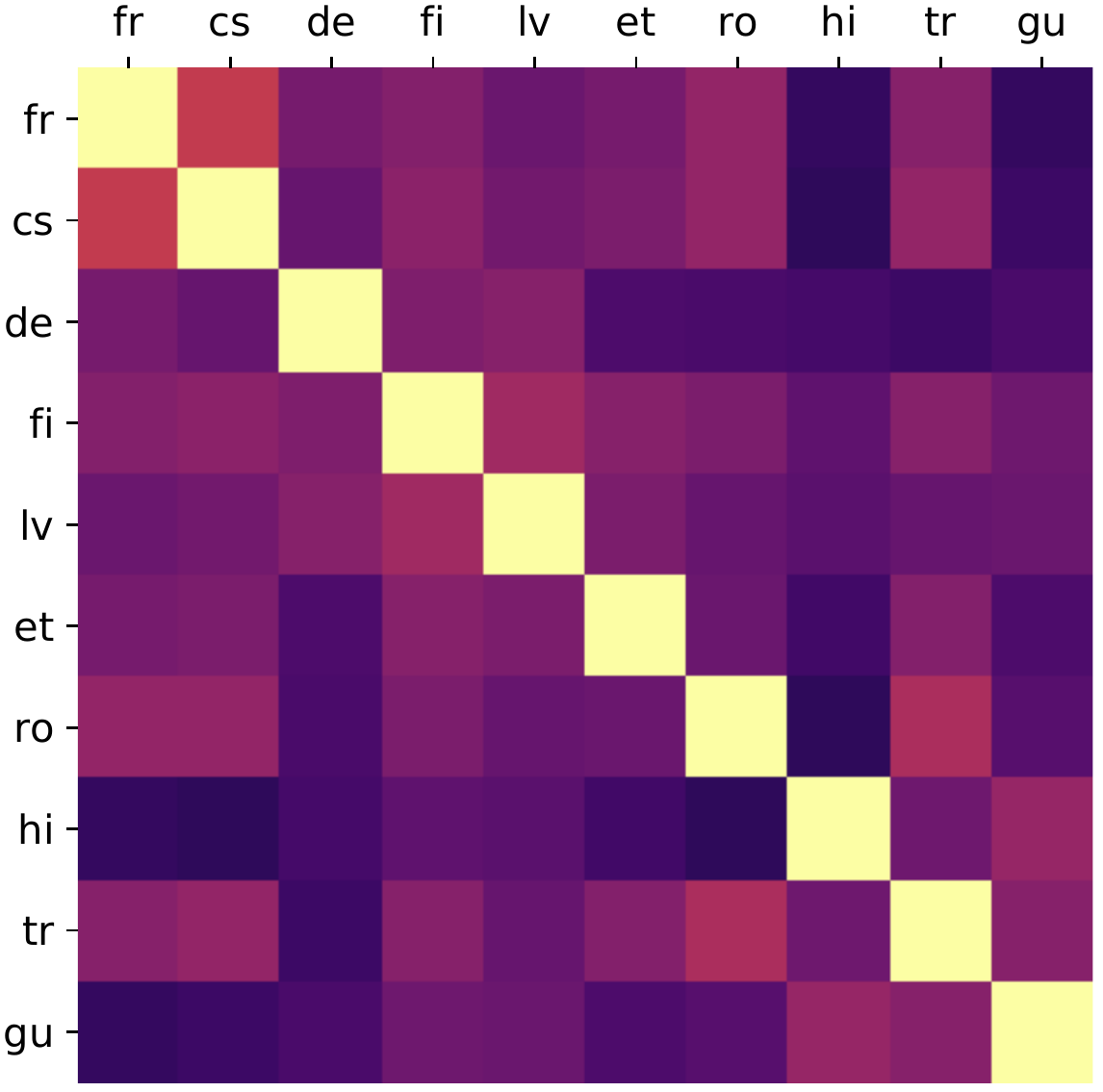}
    \label{grads_conflict_xlmt}
    }
    \caption{Cosine similarities between two gradients of training directions in (a) the baseline and (b) our method. Lower similarity (darker color) means higher negative interference.}
    % \vspace{-10pt}
\end{figure}
%%%%%%%%%%%%%%%%%%%%%%%%%%%%%%%%%%%%%%%%%%%%%%%%%%%%%%%%%%%%

\paragraph{Conflicting Gradient} To delve into the function of the language-specific module for multilingual training \cite{gradient_surgey}, we define $\Phi(L_a, L_b) = \frac{g_{L_a} \cdot g_{L_b}}{\lVert g_{L_a} \rVert \lVert g_{L_b} \rVert}$ as the cosine similarity between two task gradients $g_{L_a}$ and $g_{L_b}$, where $g_{L_a}$ and $g_{L_b}$ separately denote the gradients of the En$\to$ L$_a$ and En$\to$ L$_b$ translation direction. $\Phi(L_a, L_b)$ determines whether $g_{L_a}$ conflicts with $g_{L_b}$ by computing the cosine similarity between vectors $g_{L_a}$ and $g_{L_b}$, where the small value indicate conflicting gradients. Figure \ref{grads_conflict_xlmt} and \ref{grads_conflict_our} show the similarities of the baseline and our method between different training directions. Different training tasks of our method have similar optimization, where $\Phi(L_a, L_b)$ has larger value compared to the baseline scores. It corroborates that our language-specific training can effectively mitigate the conflicting gradients.

%%%%%%%%%%%%%%%%%%%%%%%%%%%%%%%%%%%%%%%%%%%%%%%%%%%%%%%%%%%%
\begin{figure}[t]
    \centering
    \subfigure[Selection probabilities.]{
    \includegraphics[width=0.55\columnwidth]{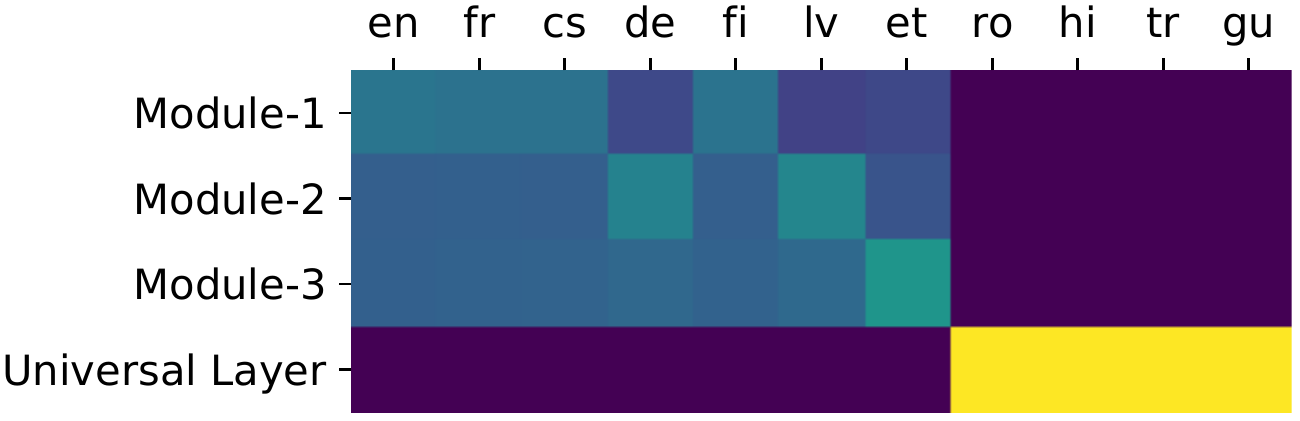}
    \label{selection_mechanism}
    }
    \subfigure[Dictionary overlaps.]{
    \includegraphics[width=0.38\columnwidth]{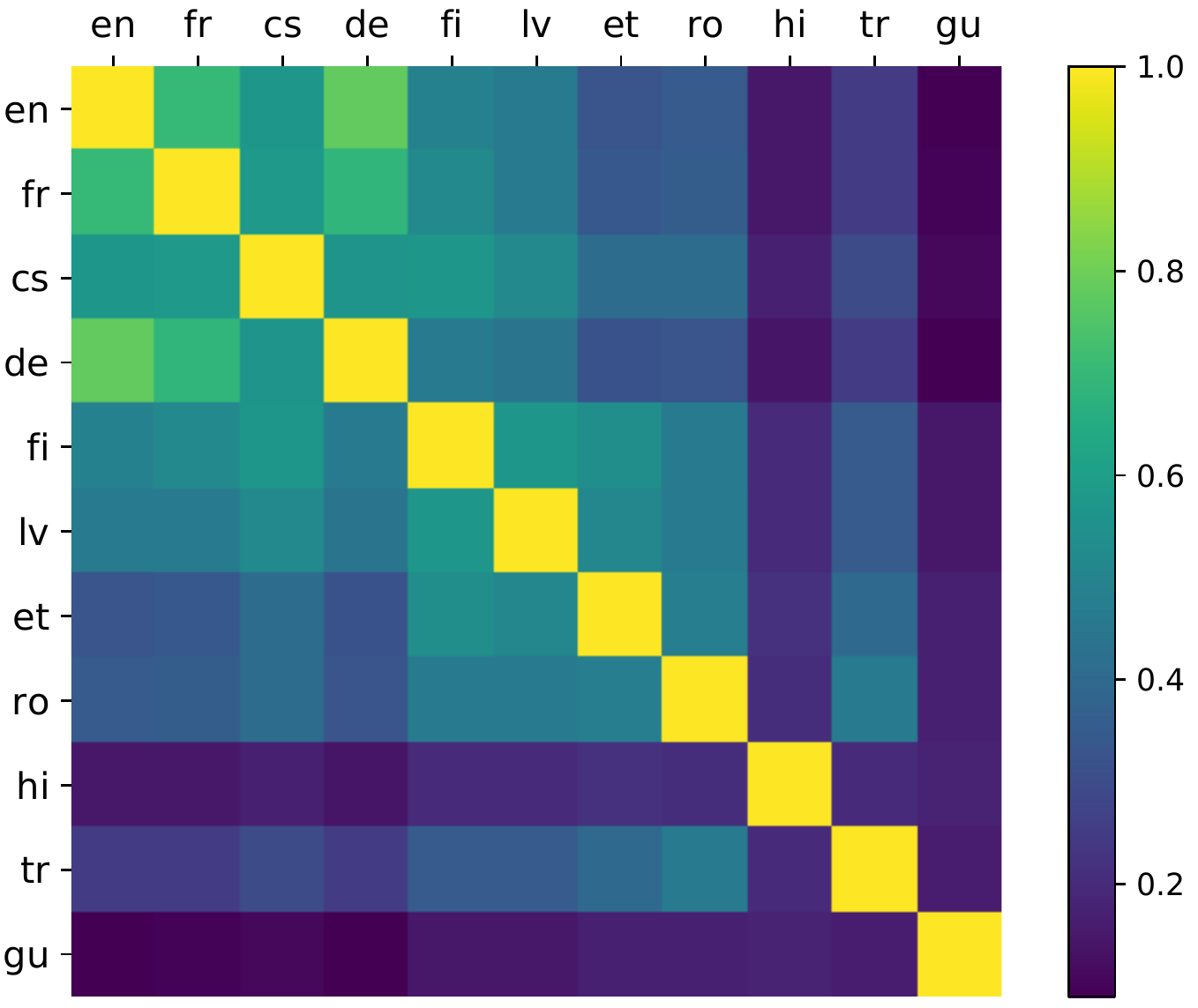}
    \label{words_overlap}
    }
    \caption{(a) is the selection probabilities of different language-specific modules generated by the embedding of the target language symbol. The universal layer is only used for the low-resource languages. (b) is the overlapping ratio of the dictionaries between different languages. Lighter green means the higher selection probability of (a) and overlapping ratio of (b).} 
    % \vspace{-10pt}
\end{figure}
%%%%%%%%%%%%%%%%%%%%%%%%%%%%%%%%%%%%%%%%%%%%%%%%%%%%%%%%%%%%

\paragraph{Language-specific Selection Mechanism} The selective language-specific pool (SLP) of high-resource languages significantly contributes to translation quality. Equation \ref{soft_adapter_probabilities} describes the selection of the module with the highest probabilities generated by the embedding of the target language symbol for the given translation direction. In Figure \ref{selection_mechanism}, $\theta_{1}$ is used for En, Fr, Cs, and Fi. $\theta_{2}$ is for De and Lv. $\theta_{3}$ is for Et. The universal layer is used for all low-resource languages.

In Figure \ref{words_overlap}, we calculate the overlaps of dictionaries among multiple languages to measure the relationship of different languages. The language similarity between $L_a$ and $L_b$ is calculated by $\text{Sim}(L_a, L_b) = \frac{\lVert \mathcal{D}_{L_a} \cap \mathcal{D}_{L_b} \rVert}{\lVert \mathcal{D}_{L_a} \cup \mathcal{D}_{L_b} \rVert}$, where $\mathcal{D}_{L_a}$ and $\mathcal{D}_{L_b}$ are the dictionary of language $L_a$ and $L_b$.
Figure \ref{words_overlap} shows that the language Et has the minimum overlap between other high-resource languages, where $\theta_{3}$ is only used for Et. Therefore, we conclude that similar languages tend to select the same language-specific module from SLP.

%%%%%%%%%%%%%%%%%%%%%%%%%%%%%%%%%%%%%%%%%%%%%%%%%%%%%%%%%%%%
\begin{figure}[t]
    \centering
    \subfigure[$2$-th]{
    \includegraphics[width=0.22\columnwidth]{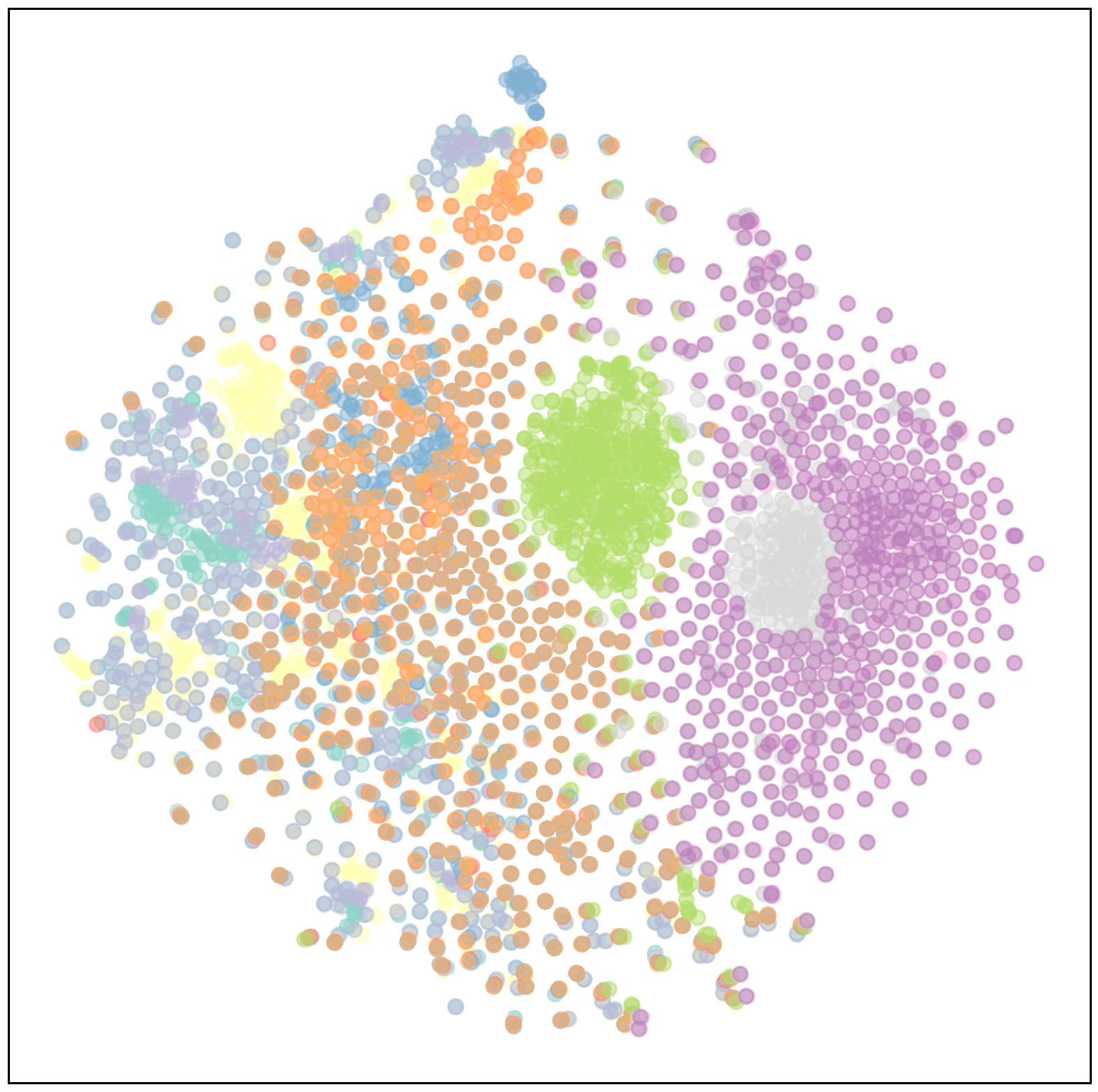}
    \label{tsne_2}
    }
    \subfigure[$3$-th]{
    \includegraphics[width=0.22\columnwidth]{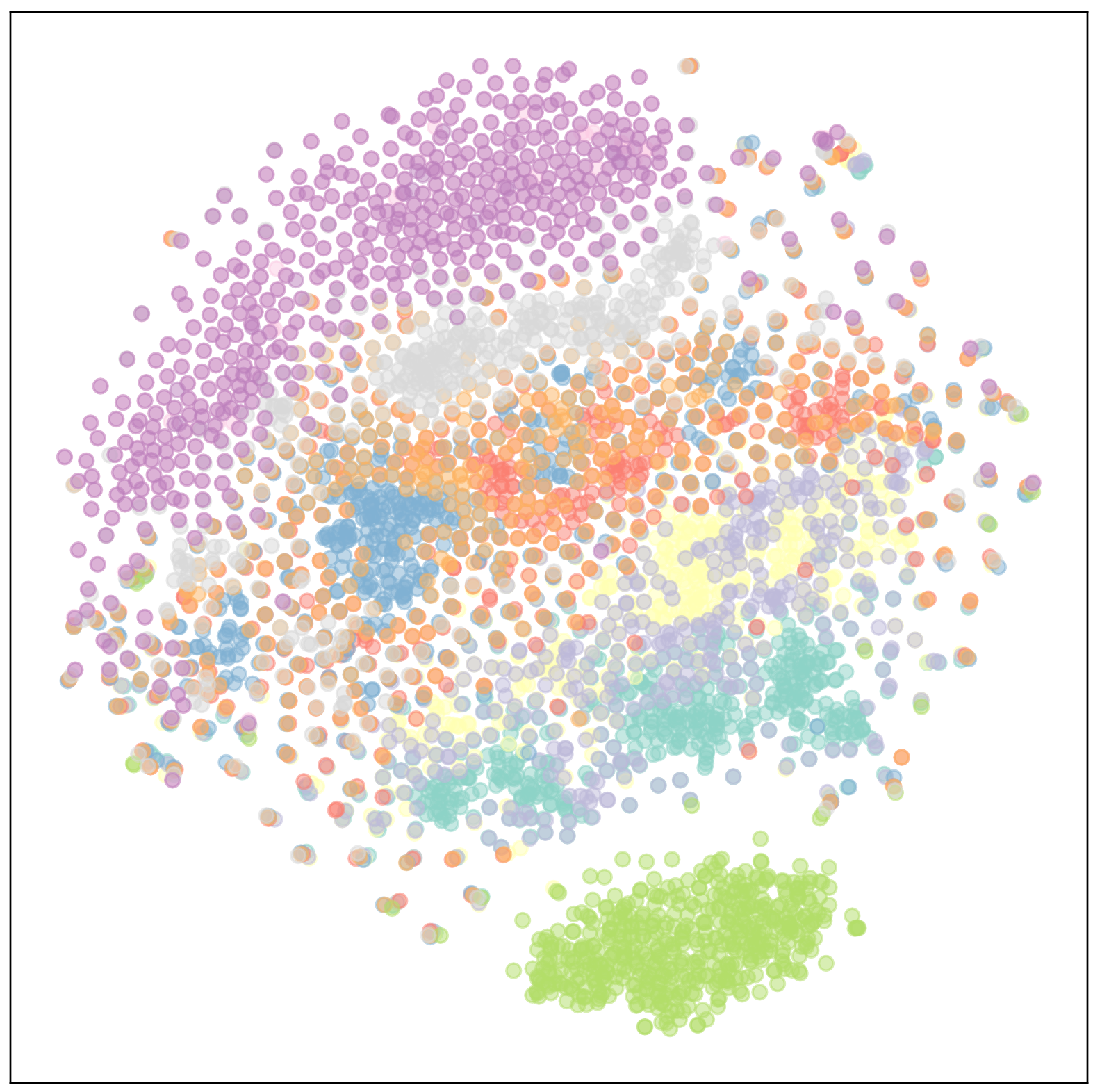}
    \label{tsne_3}
    }
    \subfigure[$6$-th]{
    \includegraphics[width=0.22\columnwidth]{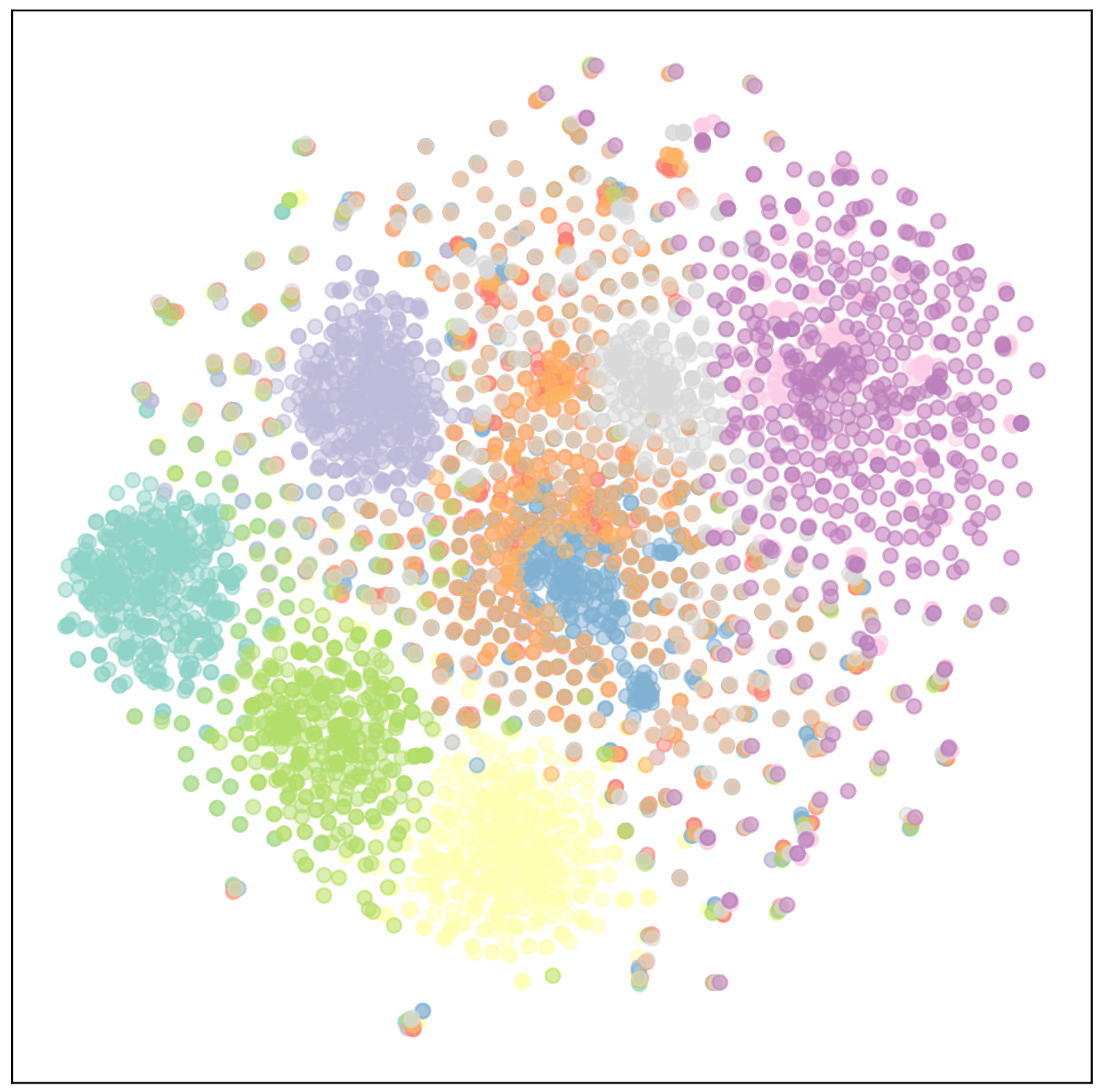}
    \label{tsne_6}
    }
    \subfigure[$7$-th]{
    \includegraphics[width=0.22\columnwidth]{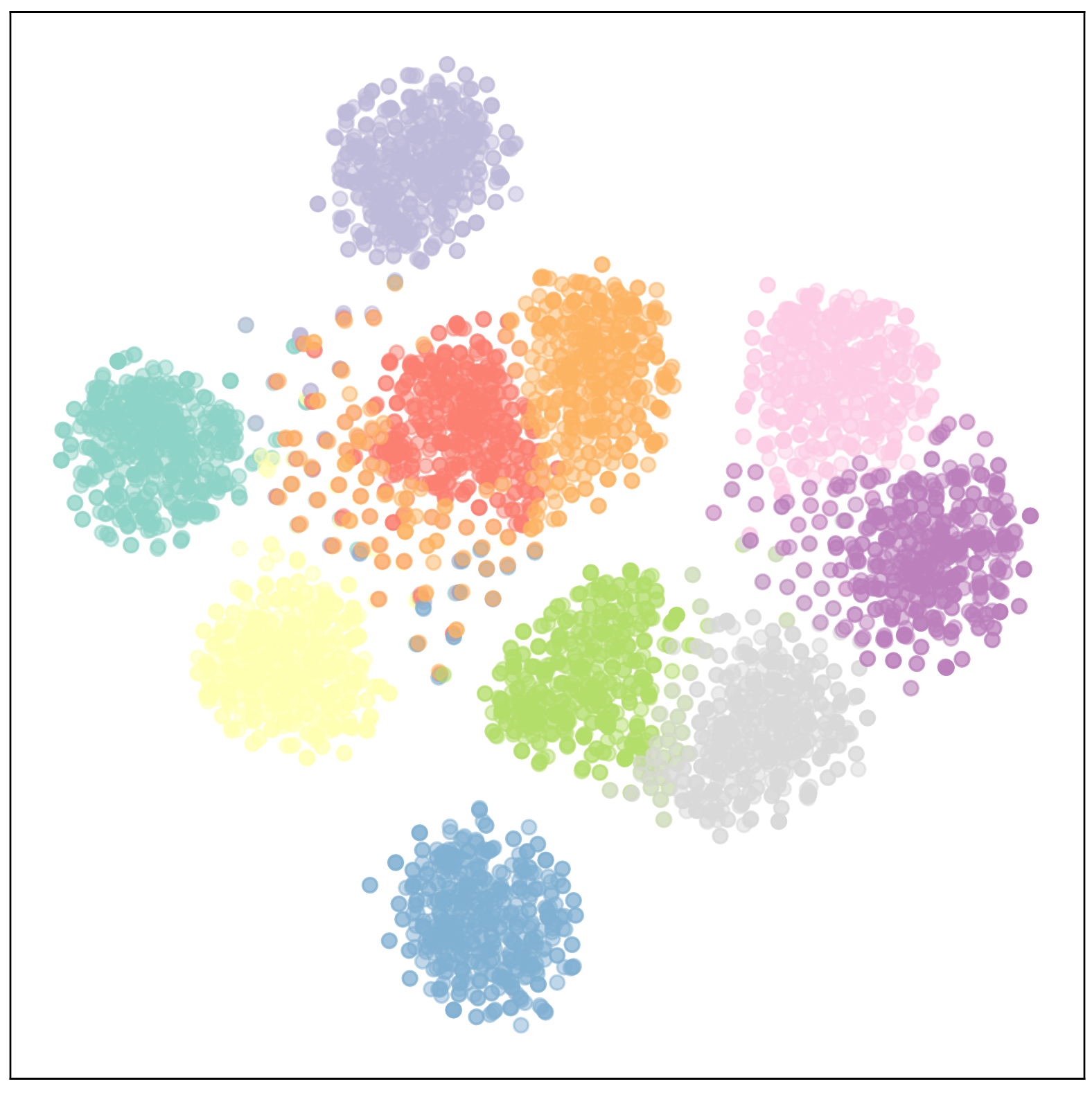}
    \label{tsne_top}
    }
    \caption{t-SNE visualization of the sentence representations from the bottom decoder layer (a), (b), (c), to the language-specific layer from SLP (d). Each color denotes one language.} 
    % \vspace{-10pt}
    \label{tsne_figures}
\end{figure}
%%%%%%%%%%%%%%%%%%%%%%%%%%%%%%%%%%%%%%%%%%%%%%%%%%%%%%%%%%%%
\paragraph{Decoder Representation Visualization} We randomly select 500 English sentences and visualize their representations \cite{t_SNE} of the bottom decoder layers and the language-specific layer in Figure \ref{tsne_figures}. The first hidden state of the decoder is regarded as the sentence representation. Compared to Figure \ref{tsne_2}, \ref{tsne_3}, and \ref{tsne_6}, different languages become more distinct and less likely to overlap with each other in Figure \ref{tsne_top}, proving that the selective language-specific pool (SLP) effectively projects the language-shared representations into language-distinct ones for better target generation of different target languages.

\section{Related Work}
Multilingual neural machine translation (MNMT) \cite{googlemnmt,mmnmt,m2m,deep_encoder_mnmt,denoising_mnmt} enables numerous translation directions by shared encoder and decoder for all languages. The MNMT system can be categorized into one-to-many \cite{one_to_many_mnmt}, many-to-one \cite{clustering_mnmt}, and many-to-many \cite{ctl_mnmt} translation. Previous studies utilize assisting high-resource languages to improve low-resource or even zero-shot translation.

While MNMT is promising, it often underperforms bilingual baselines due to the interference in shared parameters, especially on high-resource pairs \cite{negative_interference_mnmt}. To address this issue, language-specific modules are proposed to both enhance the low-resource translation and maintain the high-resource performance. Recent works mainly focus on designing language-specific components to boost the rich-resource translation quality \cite{ls_attention_mnmt,monolingual_adapter,adaptive_sparse_transformer}. Further works discuss when and where language-specific capacity matters most in MNMT \cite{close_gap_mnmt}. Our method finds a better balance between language-specific and language-agnostic models to mitigate negative interference.

\section{Conclusion}
In this work, we propose a novel multilingual translation model with the high-resource language-specific training called \ourmethod{}. The multilingual model is trained on multiple high-resource corpora with the selective language-specific pool, followed by continuing training on both high- and low-resource languages. Experimental results evaluated on WMT-10 and OPUS-100 benchmarks demonstrate that \ourmethod{} significantly outperforms all previous baselines.

\bibliographystyle{named}
\bibliography{ijcai22}

\end{document}

\maketitle
\appendix

\section{Dataset Details and Statistics}

\paragraph{WMT-10}Table~\ref{table:wmt10} lists the detailed statistics of 10 language pairs from WMT-10. We use a collection of parallel data in different languages from the WMT datasets.
The parallel data is paired with English and other 10 languages, including French (Fr), Czech (Cs), German (De), Finnish (Fi), Latvian (Lv), Estonian (Et), Romanian (Ro), Hindi (Hi), Turkish (Tr) and Gujarati (Gu). The corpora of the WMT benchmark, exclude WikiTiles, from the latest available year of each language are chosen. After removing the duplicated samples, we also limit the size of each parallel language pair data up to 10 million by randomly sampling from the whole corpus. We adopt the same valid and test sets from the WMT benchamark as the previous work \cite{zcode}.

\paragraph{OPUS-100}
OPUS-100 \cite{opus100} is an English-centric multilingual corpus covering 100 languages, which is randomly sampled from the OPUS collection. 
Table~\ref{table:opus-stats} summarizes the statistics of training, valid, and test samples for each language from OPUS-100.
The training set comprises up to one million sentence pairs per language pair, while the development and test sets contain up to 2000 parallel sentences. The whole dataset contains approximately 55 million sentence pairs. We remove 5 languages without the corresponding valid and test sets, constructing a corpus of 95 languages including English and other languages.
%%%%%%%%%%%%%%%%%%%%%%%%%%%%%%%%%%%%%%%%%%%%%%%%%%%%%%%
\begin{table}[htb]
\centering
\resizebox{1.0\columnwidth}{!}{
\begin{tabular}{ccccccc}
\toprule\toprule
Code & Language & \#Bitext & Training & Valid & Test \\
\midrule
Fr & French & 10M     & WMT15 & Newstest13 & Newstest15 \\
Cs & Czech & 10M      & WMT19 & Newstest16 & Newstest18 \\
De & German & 4.6M    & WMT19 & Newstest16 & Newstest18 \\
Fi & Finnish & 4.8M   & WMT19 & Newstest16 & Newstest18 \\
Lv & Latvian & 1.4M   & WMT17 & Newsdev17 & Newstest17 \\
Et & Estonian & 0.7M  & WMT18 & Newsdev18 & Newstest18 \\
Ro & Romanian & 0.5M  & WMT16 & Newsdev16 & Newstest16 \\
Hi & Hindi & 0.26M    & WMT14 & Newsdev14 & Newstest14 \\
Tr & Turkish & 0.18M  & WMT18 & Newstest16 & Newstest18 \\
Gu & Gujarati & 0.08M & WMT19 & Newsdev19 & Newstest19 \\
\bottomrule\bottomrule
\end{tabular}}
\caption{Statistics and sources of the training, valid, and test sets from WMT between English and other languages.}
\vspace{-10pt}
\label{table:wmt10}
\end{table}
%%%%%%%%%%%%%%%%%%%%%%%%%%%%%%%%%%%%%%%%%%%%%%%%%%%%%%%

\section{Baselines}
We compare our method against the bilingual and multilingual baselines. For the fair comparison, \textbf{Monolingual Adapter}, \textbf{LS-MNMT}, and \textbf{DEMSD} are implemented with the same backbone model as our method and initialized by the cross-lingual pretrained model XLM-R \cite{xlmr}. \textbf{BiNMT} \cite{transformer} is trained on the bilingual corpus based on the Transformer architecture. \textbf{MNMT} \cite{googlemnmt} is simultaneously trained on all available directions. The target language symbol is prefixed to the input sentence to indicate the translation direction. \textbf{mBART} \cite{mbart} is an encoder-decoder model pretrained on large-scale monolingual corpora and then is finetuned on all parallel corpora. \textbf{XLM-R} \cite{xlmr} is finetuned on multilingual corpora, where the encoder and interleaved decoder are initialized by the cross-lingual pretrained encoder XLM-R \cite{xlmr}. \textbf{Monolingual Adapter} \cite{monolingual_adapter} inserts adapter layers of different languages into the pretrained translation model and then finetunes on all corpora. \textbf{LS-MNMT} \cite{m2m} integrates the language-specific layers of all languages into the end of the Transformer decoder. \textbf{DEMSD} \cite{deep_encoder_mnmt} leverages a deep encoder and multiple shallow decoders where each shallow decoder is responsible for a disjoint subset of target languages.

\section{Model and Training Details}
We adopt 64 Tesla V100 GPUs for experiments on the WMT-10 and OPUS-100 dataset, where training \ourmethod{} costs about 36$\sim$48 hours on the OPUS-100 dataset. Table~\ref{tab:parameters} summarizes the model parameters of all baselines and our method on WMT-10 and OPUS-100. In this work, we set the number of modules $T=3$ of SLP (hidden size $d_h=3072$) for both benchmarks. For the WMT-10 benchmark, we adopt the high-resource languages (Fr, Cs, De, Fi, Lv, and Et) for the first stage and then finetune on all available corpora. For the OPUS-100 dataset, 74 languages are employed for the first stage, where the corresponding training size exceeds 100K.

\begin{table}[ht]
\centering
\resizebox{0.75\columnwidth}{!}{
\begin{tabular}{lrrr}
\toprule
\toprule
Models & \#Layers & \#Hidden & \#Params \\
\midrule
\bf WMT-10 \\
\midrule
BiNMT (large) & 6/6 & 1024 & 242M \\
BiNMT (small) & 3/3 & 256 & 10M \\
MNMT & 6/6 & 1024 &  242M \\
mBART & 12/12 & 1024 & 611M \\
XLM-R & 12/6 & 768 & 362M \\
Monolingual Adapter & 12/6 & 768 & 371M \\
LS-MNMT & 12/6 & 768 & 409M \\
MESED & 36/6 & 768 & 532M \\
\ourmethod{} & 12/6 & 768 & 381M \\
\midrule
\textbf{OPUS-100} \\ \midrule
MNMT & 12/6 & 768 & 362M \\
XLM-R & 12/6 & 768 & 362M \\
Monolingual Adapter & 12/6 & 768 & 437M \\
LS-MNMT & 12/6 & 768 & 456M \\
MESED & 36/6 & 768 & 532M \\
\ourmethod{} & 12/6 & 768 & 381M \\
\bottomrule\bottomrule
\end{tabular}}
\caption{Model sizes of all methods on two datasets. BiNMT (large) is used for high-resource directions, while BiNMT (small) is used for low-resource directions.}
\vspace{-10pt}
\label{tab:parameters}
\end{table}

\section{Results on WMT-10 and OPUS-100}
We provide the evaluation results of the multilingual baseline, XLM-R, and our method for all 94 language pairs on OPUS-100 test sets.
Table~\ref{table:opus-x2e} and Table~\ref{table:opus-e2x} report the BLEU points on X $\to$ En and En $\to$ X test sets respectively. Our method in the many-to-many setting significantly outperforms MNMT and XLM-R by a large margin. The result of our model is statistically significant compared to the baselines ($p < 0.05$) on the WMT-10 and OPUS-100 benchmark.

\begin{table*}[t]
\centering
\small
\begin{tabular}{llrrrp{1cm}llrrr}
\toprule
\toprule
Code & Language & Train & Valid & Test &  & Code & Language & Train & Valid & Test \\
af & Afrikaans & 275512 & 2000 & 2000 & & lv & Latvian & 1000000 & 2000 & 2000 \\
am & Amharic & 89027 & 2000 & 2000 & & mg & Malagasy & 590771 & 2000 & 2000 \\
ar & Arabic & 1000000 & 2000 & 2000 & & mk & Macedonian & 1000000 & 2000 & 2000 \\
as & Assamese & 138479 & 2000 & 2000 & & ml & Malayalam & 822746 & 2000 & 2000 \\
az & Azerbaijani & 262089 & 2000 & 2000 & & mr & Marathi & 27007 & 2000 & 2000 \\
be & Belarusian & 67312 & 2000 & 2000 & & ms & Malay & 1000000 & 2000 & 2000 \\
bg & Bulgarian & 1000000 & 2000 & 2000 & & mt & Maltese & 1000000 & 2000 & 2000 \\
bn & Bengali & 1000000 & 2000 & 2000 & & my & Burmese & 24594 & 2000 & 2000 \\
br & Breton & 153447 & 2000 & 2000 & & nb & Norwegian Bokmål & 142906 & 2000 & 2000 \\
bs & Bosnian & 1000000 & 2000 & 2000 & & ne & Nepali & 406381 & 2000 & 2000 \\
ca & Catalan & 1000000 & 2000 & 2000 & & nl & Dutch & 1000000 & 2000 & 2000 \\
cs & Czech & 1000000 & 2000 & 2000 & & nn & Norwegian Nynorsk & 486055 & 2000 & 2000 \\
cy & Welsh & 289521 & 2000 & 2000 & & no & Norwegian & 1000000 & 2000 & 2000 \\
da & Danish & 1000000 & 2000 & 2000 & & oc & Occitan & 35791 & 2000 & 2000 \\
de & German & 1000000 & 2000 & 2000 & & or & Oriya & 14273 & 1317 & 1318 \\
el & Greek & 1000000 & 2000 & 2000 & & pa & Panjabi & 107296 & 2000 & 2000 \\
eo & Esperanto & 337106 & 2000 & 2000 & & pl & Polish & 1000000 & 2000 & 2000 \\
es & Spanish & 1000000 & 2000 & 2000 & & ps & Pashto & 79127 & 2000 & 2000 \\
et & Estonian & 1000000 & 2000 & 2000 & & pt & Portuguese & 1000000 & 2000 & 2000 \\
eu & Basque & 1000000 & 2000 & 2000 & & ro & Romanian & 1000000 & 2000 & 2000 \\
fa & Persian & 1000000 & 2000 & 2000 & & ru & Russian & 1000000 & 2000 & 2000 \\
fi & Finnish & 1000000 & 2000 & 2000 & & rw & Kinyarwanda & 173823 & 2000 & 2000 \\
fr & French & 1000000 & 2000 & 2000 & & se & Northern Sami & 35907 & 2000 & 2000 \\
fy & Western Frisian & 54342 & 2000 & 2000 & & sh & Serbo-Croatian & 267211 & 2000 & 2000 \\
ga & Irish & 289524 & 2000 & 2000 & & si & Sinhala & 979109 & 2000 & 2000 \\
gd & Gaelic & 16316 & 1605 & 1606 & & sk & Slovak & 1000000 & 2000 & 2000 \\
gl & Galician & 515344 & 2000 & 2000 & & sl & Slovenian & 1000000 & 2000 & 2000 \\
gu & Gujarati & 318306 & 2000 & 2000 & & sq & Albanian & 1000000 & 2000 & 2000 \\
ha & Hausa & 97983 & 2000 & 2000 & & sr & Serbian & 1000000 & 2000 & 2000 \\
he & Hebrew & 1000000 & 2000 & 2000 & & sv & Swedish & 1000000 & 2000 & 2000 \\
hi & Hindi & 534319 & 2000 & 2000 & & ta & Tamil & 227014 & 2000 & 2000 \\
hr & Croatian & 1000000 & 2000 & 2000 & & te & Telugu & 64352 & 2000 & 2000 \\
hu & Hungarian & 1000000 & 2000 & 2000 & & tg & Tajik & 193882 & 2000 & 2000 \\
id & Indonesian & 1000000 & 2000 & 2000 & & th & Thai & 1000000 & 2000 & 2000 \\
ig & Igbo & 18415 & 1843 & 1843 & & tk & Turkmen & 13110 & 1852 & 1852 \\
is & Icelandic & 1000000 & 2000 & 2000 & & tr & Turkish & 1000000 & 2000 & 2000 \\
it & Italian & 1000000 & 2000 & 2000 & & tt & Tatar & 100843 & 2000 & 2000 \\
ja & Japanese & 1000000 & 2000 & 2000 & & ug & Uighur & 72170 & 2000 & 2000 \\
ka & Georgian & 377306 & 2000 & 2000 & & uk & Ukrainian & 1000000 & 2000 & 2000 \\
kk & Kazakh & 79927 & 2000 & 2000 & & ur & Urdu & 753913 & 2000 & 2000 \\
km & Central Khmer & 111483 & 2000 & 2000 & & uz & Uzbek & 173157 & 2000 & 2000 \\
kn & Kannada & 14537 & 917 & 918 & & vi & Vietnamese & 1000000 & 2000 & 2000 \\
ko & Korean & 1000000 & 2000 & 2000 & & wa & Walloon & 104496 & 2000 & 2000 \\
ku & Kurdish & 144844 & 2000 & 2000 & & xh & Xhosa  & 439671 & 2000 & 2000 \\
ky & Kyrgyz & 27215 & 2000 & 2000 & & yi & Yiddish & 15010 & 2000 & 2000 \\
li & Limburgan & 25535 & 2000 & 2000 & & zh & Chinese & 1000000 & 2000 & 2000 \\
lt & Lithuanian & 1000000 & 2000 & 2000 & & zu & Zulu & 38616 & 2000 & 2000 \\
\bottomrule
\bottomrule
\end{tabular}
\caption{Statistics of the training, valid, and test sets from OPUS-100. The languages are arranged by the alphabet order.}
\label{table:opus-stats}
\end{table*}

%%%%%%%%%%%%%%%%%%%%%%%x->en%%%%%%%%%%%%%%%%%%%%%%%%%%%%%%%%
\begin{table*}[t]
\centering
\resizebox{0.8\textwidth}{!}{
\begin{tabular}{l|cccccccccccccc}
\toprule
\toprule
Code & af & am & ar & as & az & be & bg & bn & br & bs & ca & cs & cy & da \\
\midrule
MNMT & 52.0 & 21.6 & 37.4 & 56.3 & 26.0 & 28.1 & 32.6 & 22.2 & 24.0 & 31.5 & 39.4 & 34.6 & 46.9 & 36.9 \\
XLM-R & 52.5 & 22.4 & 38.7 & 57.8 & 26.0 & 28.2 & 33.2 & 23.1 & 24.3 & 32.1 & 40.1 & 35.7 & 49.4 & 37.8 \\
\ourmethod{} & 53.7 &20.4 &39.8 &59.5 &25.0 &28.7 &34.1 &23.8 &24.0 &32.7 &41.0 &36.8 &51.3 &39.1 \\
\midrule
Code & de & el & eo & es & et & eu & fa & fi & fr & fy & ga & gd & gl & gu \\
\midrule
MNMT & 34.7 & 33.0 & 39.1 & 40.4 & 38.2 & 20.8 & 22.5 & 25.5 & 34.1 & 40.8 & 63.2 & 76.6 & 31.1 & 60.2 \\
XLM-R & 35.1 & 34.0 & 39.4 & 40.9 & 36.2 & 21.6 & 23.4 & 26.6 & 35.2 & 40.6 & 65.0 & 69.9 & 31.8 & 60.9 \\
\ourmethod{} & 36.5 &35.1 &40.5 &42.1 &39.0 &22.8 &23.5 &27.6 &36.7 &42.6 &68.9 &72.6 &33.0 &62.6 \\
\midrule
Code & ha & he & hi & hr & hu & id & ig & is & it & ja & ka & kk & km & kn \\
\midrule
MNMT & 22.4 & 35.2 & 27.1 & 31.4 & 28.4 & 34.4 & 52.3 & 24.0 & 36.1 & 14.0 & 23.8 & 29.2 & 38.4 & 38.7 \\
XLM-R & 22.4 & 37.1 & 27.0 & 33.1 & 29.6 & 35.7 & 53.8 & 24.5 & 36.9 & 15.2 & 23.7 & 30.0 & 37.3 & 41.9 \\
\ourmethod{} & 21.4 &38.1 &26.2 &33.4 &30.8 &36.0 &50.6 &26.2 &37.8 &15.5 &24.4 &28.7 &40.7 &39.6 \\
\midrule
Code & ko & ku & ky & li & lt & lv & mg & mk & ml & mr & ms & mt & my & nb \\
\midrule
MNMT & 15.3 & 24.3 & 39.7 & 40.3 & 42.8 & 46.5 & 29.0 & 35.0 & 19.2 & 50.6 & 29.8 & 63.3 & 19.1 & 44.1 \\
XLM-R & 15.8 & 23.2 & 40.0 & 39.1 & 44.5 & 47.6 & 30.0 & 35.7 & 20.5 & 52.7 & 31.3 & 64.4 & 21.0 & 46.6 \\
\ourmethod{} & 16.6 &23.4 &40.4 &34.6 &45.8 &49.4 &31.8 &37.0 &20.4 &51.1 &31.4 &65.8 &19.5 &46.8 \\
\midrule
Code & ne & nl & nn & no & oc & or & pa & pl & ps & pt & ro & ru & rw & se \\
MNMT & 47.2 & 31.8 & 36.6 & 26.1 & 16.8 & 32.9 & 43.8 & 27.3 & 36.7 & 37.0 & 38.7 & 34.6 & 28.6 & 17.2 \\
XLM-R & 47.0 & 32.1 & 37.0 & 26.8 & 18.4 & 34.9 & 46.7 & 27.8 & 37.7 & 37.8 & 39.5 & 35.8 & 28.5 & 18.3 \\
\ourmethod{} & 49.5 &33.4 &39.0 &27.5 &21.6 &33.4 &49.4 &29.5 &38.5 &39.0 &40.3 &37.0 &30.1 &22.9 \\
\midrule
Code & sh & si & sk & sl & sq & sr & sv & ta & te & tg & th & tk & tr & tt \\
\midrule
MNMT & 56.6 & 23.0 & 39.1 & 28.8 & 42.7 & 31.2 & 32.9 & 28.3 & 42.9 & 23.5 & 22.3 & 46.6 & 25.1 & 19.1 \\
XLM-R & 57.4 & 24.3 & 39.9 & 29.5 & 43.6 & 32.1 & 33.1 & 28.9 & 45.9 & 23.9 & 22.7 & 46.8 & 26.2 & 19.3 \\
\ourmethod{} & 60.7 &25.4 &41.0 &30.8 &44.7 &33.3 &33.7 &28.7 &45.1 &23.5 &23.7 &47.2 &26.9 &17.8 \\
\midrule
Code & ug & uk & ur & uz & vi & wa & xh & yi & zh & zu &  &  &  &  \\
\midrule
MNMT & 19.3 & 27.7 & 19.7 & 18.8 & 25.8 & 32.1 & 26.0 & 32.5 & 38.8 & 47.8 &  &  &  &  \\
XLM-R & 18.9 & 28.4 & 19.8 & 18.0 & 27.0 & 29.3 & 27.6 & 25.8 & 39.7 & 48.9 &  &  &  &  \\
\ourmethod{} & 17.8 &29.6 &17.8 &18.0 &27.2 &31.6 &30.2 &26.5 &41.4 &46.9 \\
\bottomrule
\bottomrule
\end{tabular}}
\caption{X $\rightarrow$ En evaluation results on test sets for 94 language pairs in many-to-many setting on the OPUS-100 test sets.}
\label{table:opus-x2e}
\end{table*}
%%%%%%%%%%%%%%%%%%%%%%%%%%%%%%%%%%%%%%%%%%%%%%%%%%%%%%%

%%%%%%%%%%%%%%%%%%%%%%%en->x%%%%%%%%%%%%%%%%%%%%%%%%%%%%%%%%
\begin{table*}[t]
\centering
\resizebox{0.8\textwidth}{!}{
\begin{tabular}{l|cccccccccccccc}
\toprule
\toprule
Code & af & am & ar & as & az & be & bg & bn & br & bs & ca & cs & cy & da \\
\midrule
MNMT & 45.7 & 20.8 & 20.5 & 42.4 & 28.5 & 26.6 & 29.6 & 11.9 & 25.6 & 21.9 & 36.1 & 26.9 & 43.5 & 35.6 \\
XLM-R & 47.0 & 21.1 & 21.4 & 42.0 & 29.3 & 28.2 & 30.2 & 12.2 & 26.0 & 22.7 & 37.0 & 28.3 & 44.1 & 36.2 \\
\ourmethod{} & 48.1 &22.9 &22.7 &43.5 &30.2 &27.9 &31.4 &12.5 &27.9 &23.2 &38.4 &29.0 &45.3 &37.1 \\
\midrule
Code & de & el & eo & es & et & eu & fa & fi & fr & fy & ga & gd & gl & gu \\
\midrule
MNMT & 30.3 & 27.6 & 34.9 & 37.7 & 32.3 & 14.9 & 10.0 & 21.4 & 32.9 & 34.4 & 51.7 & 31.7 & 27.9 & 53.5 \\
XLM-R & 31.4 & 27.8 & 35.0 & 38.1 & 29.0 & 15.2 & 10.2 & 22.6 & 33.8 & 34.2 & 52.8 & 30.2 & 28.4 & 54.1 \\
\ourmethod{} & 31.5 &28.9 &36.0 &38.7 &32.0 &16.2 &11.3 &24.1 &35.0 &34.5 &56.7 &54.9 &29.5 &54.8 \\
\midrule
Code & ha & he & hi & hr & hu & id & ig & is & it & ja & ka & kk & km & kn \\
\midrule
MNMT & 51.3 & 29.4 & 19.7 & 24.3 & 20.5 & 29.8 & 45.7 & 22.0 & 30.2 & 12.2 & 17.9 & 25.1 & 19.4 & 28.8 \\
XLM-R & 53.0 & 29.0 & 19.6 & 24.9 & 21.7 & 29.8 & 45.8 & 22.8 & 31.5 & 12.9 & 18.7 & 25.5 & 19.6 & 29.8 \\
\ourmethod{} & 54.0 &31.1 &17.8 &25.9 &22.9 &30.9 &43.6 &24.2 &32.5 &13.2 &19.2 &25.8 &19.5 &28.1 \\
\midrule
Code & ko & ku & ky & li & lt & lv & mg & mk & ml & mr & ms & mt & my & nb \\
\midrule
MNMT & 6.2 & 8.2 & 34.4 & 32.5 & 36.3 & 40.3 & 23.0 & 33.7 & 6.1 & 33.7 & 24.8 & 48.5 & 11.4 & 38.7 \\
XLM-R & 6.8 & 8.4 & 34.4 & 33.3 & 36.9 & 41.2 & 23.5 & 34.6 & 5.8 & 34.2 & 24.9 & 49.0 & 11.8 & 39.1 \\
\ourmethod{} & 7.1 &8.5 &32.8 &33.3 &38.8 &42.8 &24.7 &36.2 &6.2 &33.9 &26.0 &49.9 &10.6 &40.2 \\
\midrule
Code & ne & nl & nn & no & oc & or & pa & pl & ps & pt & ro & ru & rw & se \\
\midrule
MNMT & 43.4 & 28.1 & 31.5 & 28.8 & 24.3 & 34.0 & 43.7 & 21.3 & 43.1 & 32.3 & 31.4 & 29.1 & 67.9 & 26.0 \\
XLM-R & 42.2 & 28.8 & 31.6 & 29.4 & 23.5 & 33.0 & 44.6 & 22.0 & 43.7 & 33.2 & 31.7 & 29.9 & 68.5 & 27.7 \\
\ourmethod{} & 43.8 &29.6 &33.2 &30.2 &25.3 &33.8 &44.5 &23.8 &41.4 &34.3 &33.4 &31.6 &68.1 &25.1 \\
\midrule
Code & sh & si & sk & sl & sq & sr & sv & ta & te & tg & th & tk & tr & tt \\
\midrule
MNMT & 52.0 & 10.8 & 30.5 & 24.8 & 38.3 & 21.8 & 32.0 & 19.8 & 32.8 & 32.0 & 9.8 & 43.6 & 17.5 & 23.6 \\
XLM-R & 53.4 & 11.0 & 30.9 & 25.4 & 39.0 & 21.8 & 32.7 & 19.7 & 33.4 & 33.0 & 10.6 & 45.7 & 17.2 & 24.7 \\
\ourmethod{} & 55.9 &11.4 &32.7 &26.9 &39.9 &22.9 &33.4 &20.2 &32.4 &40.9 &9.0 &43.4 &19.0 &31.7\\
\midrule
Code & ug & uk & ur & uz & vi & wa & xh & yi & zh & zu &  &  &  &  \\
\midrule
MNMT & 12.6 & 16.8 & 19.0 & 17.6 & 22.4 & 27.7 & 14.3 & 27.2 & 41.9 & 35.2 &  &  &  &  \\
XLM-R & 13.1 & 17.1 & 18.9 & 18.8 & 23.1 & 29.0 & 14.7 & 28.4 & 42.4 & 36.1 &  &  &  &  \\
\ourmethod{} & 12.8 &19.2 &16.5 &23.0 &23.9 &29.2 &16.6 &28.3 &43.4 &36.1 \\
\bottomrule
\bottomrule
\end{tabular}}
\caption{En $\rightarrow$ X evaluation results on test sets for 94 language pairs in many-to-many setting on the OPUS-100 test sets.}
\label{table:opus-e2x}
\end{table*}
%%%%%%%%%%%%%%%%%%%%%%%%%%%%%%%%%%%%%%%%%%%%%%%%%%%%%%%

\section{Training Details and Evaluation}
We adopt the Transformer architecture as the backbone model for all our experiments, which has 12 encoder and 6 decoder layers with an embedding size of 768, a dropout of 0.1, the feed-forward network size of 3072, and 12 attention heads. We use the XLM-R to initialize the Transformer model as the previous work \cite{xlmt,deltalm}, where the encoder and interleaved decoder are initialized by the cross-lingual pretrained encoder XLM-R \cite{xlmr}. For fair comparison, other baselines are also initialized in the same way.
We train multilingual models with Adam \cite{adam} ($\beta_{1}=0.9$, $\beta_{2}=0.98$). The learning rate is set as 5e-4 with a warm-up step of 4,000. The models are trained with the label smoothing cross-entropy with a smoothing ratio of 0.1. The batch size is 4096 tokens on 64 Tesla V100 GPUs. For WMT-10, we first train the multilingual model with 6 languages and then finetunes on all languages. For OPUS-100, the model is trained in the languages where the number of pairs exceeds 10K. 

To mitigate the unbalance of the multiple bilingual corpora, \cite{mnmt_challenges,zcode}, we employ the temperature-based sampling method, where the temperature gradually increases to the peak value for several epochs. The temperature is calculated by $\tau_{i}=\min(\tau, \tau_{0}+\frac{i}{N}(\tau-\tau_{0}))$,
where $\tau_0$ and $\tau$ separately denote the initial and peak temperature, and $N$ is the number of warming-up epochs. For a fair comparison, we set $\tau_0=1.0$, $\tau=5.0$, and $N=5$ for all experiments. We average the last 5 checkpoints and employ the beam search strategy with a beam size of 5 for evaluation. The evaluation metric is the case-sensitive detokenized sacreBLEU\footnote{BLEU+case.mixed+lang.\{src\}-\{tgt\}+numrefs.1+smooth.exp+tok.13a+version.1.4.14} \cite{sacrebleu}.

\section{Time Cost of Two-Stage Training} 
All baselines implemented by ourselves are trained for 20 epochs (nearly 24 hours) to converge. The first stage of our method is trained for 12$\sim$18 epochs (14$\sim$20 hours) and the second stage requires only 4$\sim$6 (6$\sim$8 hours) epochs for convergence. All methods are trained with 64 Tesla V100 GPUs on WMT-10.
Therefore, our method has similar training time to all baselines.

\paragraph{Effect of Disparity Loss} To encourage different languages to select different language-specific modules of SLP, we minimize the disparity loss $L_d$, which measures the similarity of language-specific layer selection among languages. The disparity loss can be regarded as an auxiliary task to encourage our method to select distinct modules for sentences in different languages. Specifically, we normalize the disparity loss by the target tokens to ensure that the optimization for disparity loss will not hurt the model performance.

\paragraph{Performance on the low-resource languages} Compared to the baseline (17.8 BLEU points), our method (18.8 BLEU points) still gets consistent improvement on the low-resource languages. We find that the two-stage training brings improvement for high-resource languages (+0.6 BLEU points) and low-resource languages (+0.3 BLEU points).
Furthermore, our method gets a stable improvement over all translation directions. The result of our model is statistically significant compared to the baselines (p $<$ 0.05) on all benchmarks.

\paragraph{More SLP Modules} On the OPUS-100 benchmark, we conduct the analytic experiments by increasing the number of SLP modules and observe that more SLP modules lead to better performance. But when the number of SLP modules reaches a certain value, the model performance has little improvement. Our model with a very large value of SLP modules ($T=20$) outperforms the model ($T=3$) by +0.8 BLEU points.
Considering the trade-off between the model performance and computational resources, we decide the proper number of SLP modules according to the performance on the validation set.

\section{Related Work}

\paragraph{Multilingual Neural Machine Translation}
Multilingual neural machine translation (MNMT) model \cite{googlemnmt,mmnmt,m2m,deep_encoder_mnmt,bimt_to_mnmt,denoising_mnmt} enables numerous translation directions by a shared encoder and decoder for all languages. The MNMT system can be categorized into one-to-many \cite{one_to_many_mnmt}, many-to-one \cite{clustering_mnmt}, and many-to-many \cite{unsupervised_mnmt,distillation_mnmt,ctl_mnmt,multilingual_agreement} translation using multiple parallel corpora. Previous studies have explored the approaches that utilize assisting high-resource languages to improve low-resource or even zero-shot translation without direct corpora by transfer learning \cite{monolingual_adapter,adapter_1600}. Different from previous methods, our method leverages the language-specific parameter pool (SLP) to pretrain the multilingual translation model on high-resource pairs and then transfers the knowledge from high-resource pairs to low-resource pairs.

\paragraph{Language-Specific Module}
While MNMT is promising, it often underperforms the bilingual baseline caused by the interference in the shared parameters, especially on the high-resource pairs \cite{mnmt_survey}. To mitigate this problem, language-specific modules are proposed to enhance the performance of low-resource languages and maintain the performance of high-resource languages. Recent efforts mainly focus on designing language-specific components for multilingual NMT to boost the translation quality on rich-resource languages, such as language-specific encoder, decoder, or attention modules \cite{ls_decoder,ls_attention_mnmt,ls_modularized,monolingual_adapter,ls_subnet,adaptive_sparse_transformer,adapter_1600,recipes_adapter_mnmt}. The recent works \cite{share_or_not,close_gap_mnmt} have discussed when and where language-specific capacity matters most in MNMT. To mitigate the negative interference introduced by multiple training directions, our method aims to find a better balance between language-specific and language-agnostic representations via the high-resource language-specific training.

\bibliographystyle{named}
\bibliography{ijcai22}